\documentclass[lettersize,journal,compsoc]{IEEEJtran}
\ifCLASSOPTIONcompsoc
  \usepackage[nocompress]{cite}
\else
  \usepackage{cite}
\fi
\ifCLASSINFOpdf
\else
\fi
\usepackage{amsmath, xspace, comment}
\usepackage{tikz}
\usepackage{algpseudocode}
\usepackage[linesnumbered]{algorithm2e}
\usepackage{colortbl}
\usepackage{multirow}
\usepackage{textcase} 
\usepackage{enumitem}
\usepackage{balance}
\usepackage{graphicx}
\usepackage{hhline}
\usepackage{xspace}
\usepackage{xcolor}
\usepackage{soul}
\usepackage{mathtools}
\usepackage{tcolorbox}
\usepackage{subcaption}
\usepackage{amssymb}
\usepackage{amsmath}
\usepackage{amsfonts}
\usepackage[hidelinks]{hyperref}
\usepackage{booktabs}
\usepackage{textcomp}
\usepackage{stfloats}
\usepackage{url}
\usepackage{verbatim}
\usepackage{cite}
\usepackage{orcidlink}
\def\BibTeX{{\rm B\kern-.05em{\sc i\kern-.025em b}\kern-.08em
    T\kern-.1667em\lower.7ex\hbox{E}\kern-.125emX}}

\newcommand*\cib[1]{\tikz[baseline=(char.base)]{
                            \node[shape=circle,fill=black,text=white,draw,inner sep=0.3pt] (char) {#1};}}
                            
\newcommand{\etal}{{\em et al.}\xspace}
\newcommand{\eg}{{\em e.g.,}\xspace}
\newcommand{\ie}{{\em i.e.,}\xspace}
\newcommand{\etc}{{\em etc.}\xspace}
\newcommand{\BfPara}[1]{{\vspace{0.5ex}\noindent\bf#1.}\xspace}

\newcommand{\ours}{{\NoCaseChange{Single\textbf{ADV}}}\xspace}

\usepackage{tcolorbox}
\tcbuselibrary{theorems}

\newtcbtheorem[]{observation}{\textbf{Observation}}
{colback=white,colframe=white!65!black,fonttitle=\bfseries,coltitle=black,boxsep = 1pt, left = 2pt, right = 2pt, top = 2pt, bottom = 2pt,
    boxrule = 1pt, bottomrule = 1pt, toprule = 1pt}{th}{}
    
\hypersetup{
	colorlinks=true,
	urlcolor=blue!70!black,
	linkcolor=red!70!black,
	citecolor=red!70!black,
} 
\setul{0.4pt}{}

\begin{document}

\title{Single-Class Target-Specific Attack against Interpretable Deep Learning Systems}

\author{Eldor~Abdukhamidov ,
        Mohammed~Abuhamad,
        George~K.~Thiruvathukal,
        Hyoungshick~Kim,
        and~Tamer~Abuhmed~
        
        \IEEEcompsocitemizethanks{\IEEEcompsocthanksitem Eldor Abdukhamidov, Hyoungshick~Kim, and Tamer ABUHMED are with the Department of Computer Science and Engineering, Sungkyunkwan University, Suwon, South Korea.\protect
        (E-mail: abdukhamidov@skku.edu, hyoung@skku.edu, tamer@skku.edu). Mohammed Abuhamad and George K. Thiruvathukal are with the Department of Computer Science, Loyola University, Chicago, USA.\protect
        (E-mail: mabuhamad@luc.edu, gkt@cs.luc.edu).\\     }
}

\markboth{Journal of \LaTeX\ Class Files,~Vol.~14, No.~8, August~2021}%
{Shell \MakeLowercase{\textit{et al.}}: A Sample Article Using IEEEtran.cls for IEEE Journals}


\IEEEtitleabstractindextext{%
\begin{abstract}
Developing a custom model specifically tailored to address domain-specific problems is crucial for achieving optimal performance. The interpretation of machine learning models plays a pivotal role in this development, aiding domain experts in gaining insights into the internal mechanisms of these models. However, adversarial attacks pose a significant threat to public trust by making interpretations of deep learning models confusing and difficult to understand. In this paper, we present a novel \textbf{\ul{Single}}-class target-specific \textbf{\ul{ADV}}ersarial attack called \ours{}. The goal of \ours{} is to generate a universal perturbation that deceives the target model into confusing a specific category of objects with a target category while ensuring highly relevant and accurate interpretations. 
The universal perturbation is stochastically and iteratively optimized by minimizing the adversarial loss that is designed to consider both the classifier and interpreter costs in targeted and non-targeted categories. In this optimization framework, ruled by the first- and second-moment estimations, the desired loss surface promotes high confidence and interpretation score of adversarial samples. By avoiding unintended misclassification of samples from other categories, \ours{} enables more effective targeted attacks on interpretable deep learning systems in both white-box and black-box scenarios. To evaluate the effectiveness of \ours{}, we conduct experiments using four different model architectures (ResNet-50, VGG-16, DenseNet-169, and Inception-V3) coupled with three interpretation models (CAM, Grad, and MASK). Through extensive empirical evaluation, we demonstrate that \ours{} effectively deceives the target deep learning models and their associated interpreters under various conditions and settings. Our experimental results show that the performance of \ours{} is effective, with an average fooling ratio of 0.74 and an adversarial confidence level of 0.78 in generating deceptive adversarial samples. Furthermore, we discuss several countermeasures against \ours{}, including a transfer-based learning approach and existing preprocessing defenses.
\end{abstract}

}
\maketitle

\IEEEdisplaynontitleabstractindextext
\IEEEpeerreviewmaketitle
\IEEEraisesectionheading{
\section{Introduction}}
\label{sec:intro}

Deep learning (DL) has made significant contributions and advancements across various domains, including computer vision \cite{ILSVRC15, simonyan2014very, he2016deep}, natural language processing, and numerous security-sensitive applications \cite{he2020towards, kindermans2019reliability}. The impressive performance of deep learning models on large datasets has gained significant attention from the research community. However, a fundamental challenge lies in comprehending the underlying factors that drive the outcomes of DL models, primarily due to their complex architectures. Consequently, converting the behavior of DL models into a more comprehensible format for end-users has become crucial. To address this issue, numerous interpretation techniques \cite{zhou2016learning, kindermans2019reliability, ILSVRC15, zhang2018interpretable} have been developed to make DL models more understandable. These models provide insight into how DL models make decisions, which can help users to trust and use these models more effectively.



DL interpretability plays a crucial role in understanding the behavior of models and ensuring confidence in detecting adversarial inputs. Thus, Interpretable Deep Learning Systems (IDLSes) have gained considerable attention, as they provide predictions and interpretations. However, recent studies have demonstrated the feasibility and practicality of creating adversarial examples (AE) that deceive both the target prediction model and its associated interpreters \cite{10.1007/978-3-030-91434-9_9, zhang2020interpretable, akhtar2021explain,juraev2022depth,abdukhamidov2022interpretations}. Consequently, IDLSes cannot guarantee robust security measures for detecting AEs.


We introduce \ours{}, a single-class target-specific adversarial attack method designed to generate targeted perturbations that deceive IDLSes by causing misclassifications of an entire class of objects (referred to as the ``source class'') into a specific ``target class.'' The perturbations are designed to maintain interpretations similar to those of benign inputs, making it difficult for IDLSes to detect them.

\begin{figure*}[t]
    \centering
    \captionsetup{justification=justified}
    \includegraphics[width=0.75\textwidth]{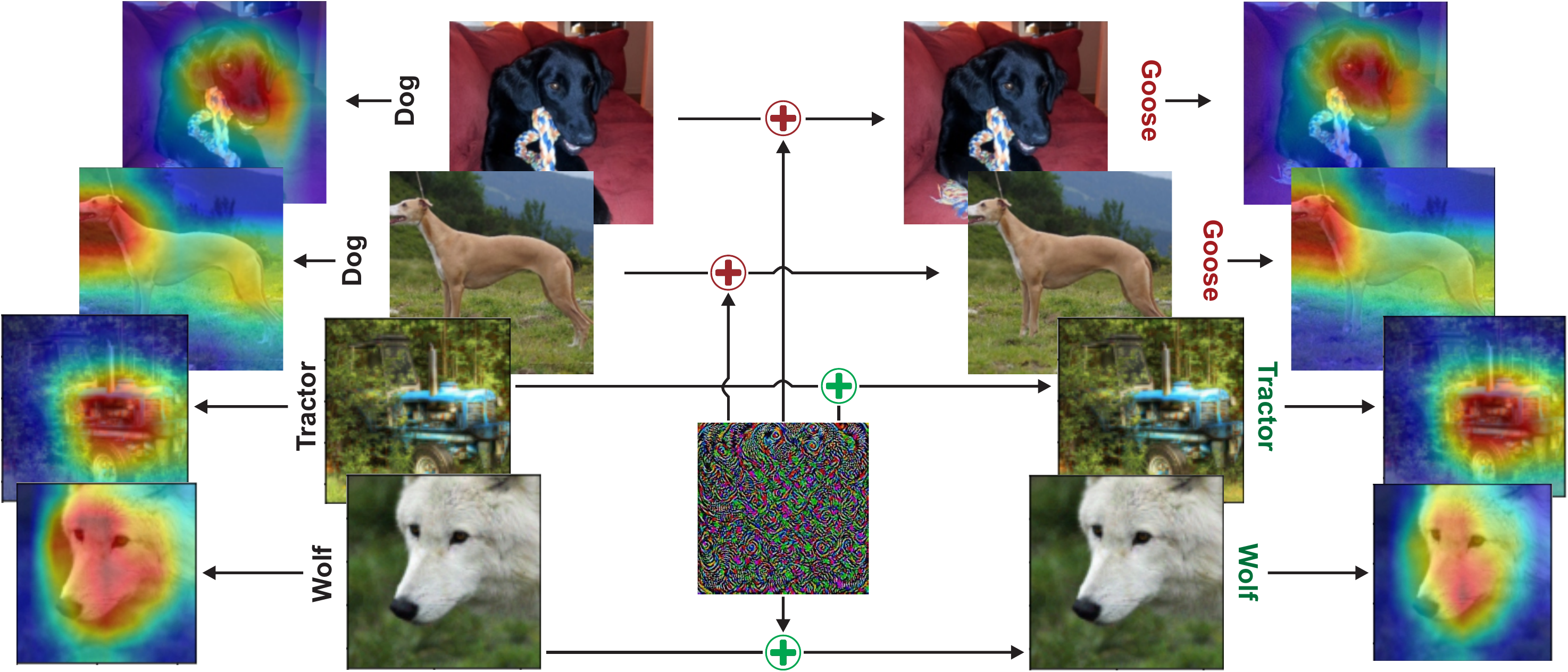}
    \caption{Adversarial example that was generated using the ResNet-50 model with the CAM interpreter. The original image is a dog, but the adversarial example is misclassified as a goose. The other class images (tractor and wolf) are not impacted. The attacker achieved this by adding a single perturbation to the original image. The perturbation was carefully chosen to have a minimal impact on the interpretation of the image while still causing the model to misclassify it.}
    \label{fig:overview}
    \vspace{-2ex}
\end{figure*}

\begin{figure*}[t]
    \centering
    \captionsetup{justification=justified}
    \includegraphics[width=0.75\textwidth]{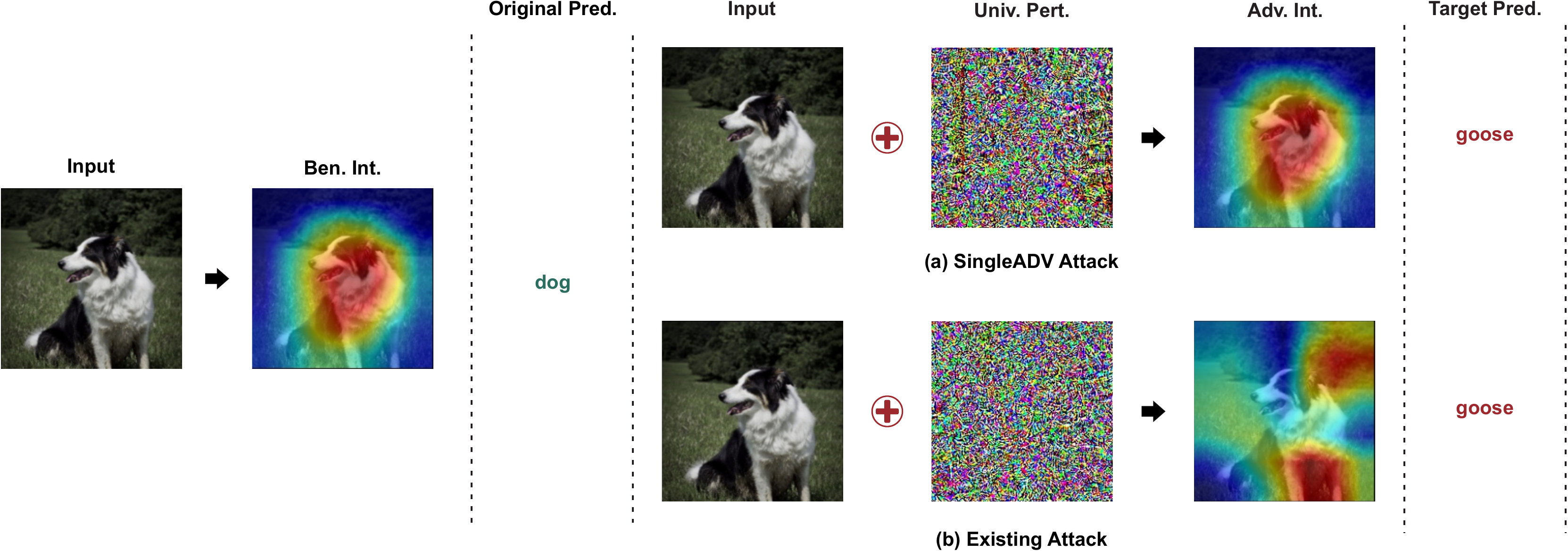}
    \caption{\ours vs. the existing attack \cite{akhtar2021explain} against ResNet-50 with CAM interpreter. \ours preserves benign attribution maps. Pred., Univ. Pert. and Adv. Int. stand for prediction, universal perturbation, and adversarial interpretation.}
    \label{fig:comparison}
    \vspace{-2ex}
\end{figure*}

In a targeted attack threat model, \ours{} generates stealthy perturbations that effectively deceive IDLSes, thereby hindering human involvement in analyzing the interpretation of adversarial inputs, as shown in \autoref{fig:overview}. Additionally, \autoref{fig:comparison} demonstrates that the employed interpreter can detect previous attacks, even in white-box scenarios. For example, \autoref{fig:overview} shows that the existing universal perturbation attacks \cite{akhtar2021explain} can be detected due to the inconsistencies between the adversarial interpretations and the object of the image, which can be recognized by an ``observer.'' On the other hand, our approach generates adversarial interpretations that are indistinguishable from benign ones, suggesting no manipulation of the input data.

The main objective of \ours is to increase the success rate of adversarial attacks. This involves fooling the classifier and misleading its interpreter while reducing the likelihood of adversarial detection. \ours achieves this by generating perturbations in a fine-grained manner. This ensures that the perturbations impact the target category while minimizing their influence on other categories.

Instead of employing the traditional targeted attack approach, where the model predicts the same label for all images, this work focuses on a single class in an adversarial scenario to reduce suspicion of an attack. We call this attack ``\textit{single-class attack}.'' Although we apply this technique to an image classification task in this paper, it can be used in various security-sensitive applications, such as malware detection and facial recognition systems. Additionally, the generated adversarial samples cause the interpreter to produce false positives, resulting in interpretations similar to benign inputs. This aspect makes it challenging to identify the involvement of the adversary. This work investigates the effects of generating universal perturbations to launch a single-class attack in both white-box and black-box scenarios. Furthermore, it explores potential countermeasures and defenses against such attacks.

\BfPara{Our Contribution} We present \ours, a single-class attack method designed to generate target-specific perturbations for inputs to fool the target deep learning models and deceive their coupled interpreters. Our method enables targeted attacks that are specific to a particular category. We evaluate the effectiveness of \ours{} on both the prediction and interpretation models using the ImageNet dataset \cite{ILSVRC15}. Our contributions can be summarized as follows:



\begin{itemize}[leftmargin=1em]
    \item We propose a novel adversarial attack method called \ours, which leverages interpretation-derived techniques to perform targeted and category-specific fooling of DL models and their associated interpreters. Unlike traditional approaches focusing on individual input samples, our method extends the scope of adversarial effects to encompass an entire object category, thus limiting the impact within the chosen category.
    \item We evaluate \ours in terms of the success rate for fooling the prediction model, the Intersection-over-Union (IoU) score for deceiving the coupled interpreter, and the leakage rate to measure the attack's impact on non-targeted classes. We demonstrate the effectiveness of our approach, \eg an average fooling ratio of 0.74 and a corresponding adversarial confidence level of 0.78.
    \item We conducted experiments using a knowledge distillation (teacher-student) approach to assess the practicality and effectiveness of \ours{} in a black-box setting. The results of our experiments confirm that \ours{} can be effectively applied in the black-box scenario, highlighting its practicality and effectiveness.
    \item We analyze the effectiveness of existing general defense techniques to mitigate \ours. Additionally, we propose a novel adversarial training method that utilizes interpretation-derived information to enhance the robustness of DL models against adversarial attacks. This approach significantly improves the model's performance when faced with adversarial examples while maintaining its performance on benign examples. 
\end{itemize}


\BfPara{Organization} The rest of the paper is organized as follows:
\autoref{sec:related} highlights the relevant literature; \autoref{sec:methods} provides the fundamental concepts, description of the problem formulation, and the main algorithm of the attack; \autoref{sec:evaluation} and \autoref{sec:blackbox} provide the experiments and results in white-box and black-box settings, respectively; \autoref{sec:countermeasures} proposes potential countermeasures; \autoref{sec:limiation} discusses the limitations; and \autoref{sec:conc} offers the conclusion.

\begin{table*}[ht]
\centering
\caption{Comparison of related works based on several aspects.}
\label{tab:comparison}
\resizebox{0.9\linewidth}{!}{%
\begin{tabular}{l>{\centering\hspace{0pt}}m{0.09\linewidth}>{\centering\hspace{0pt}}m{0.087\linewidth}>{\centering\hspace{0pt}}m{0.087\linewidth}>{\centering\hspace{0pt}}m{0.11\linewidth}>{\centering\hspace{0pt}}m{0.11\linewidth}>{\centering\hspace{0pt}}m{0.087\linewidth}>{\centering\hspace{0pt}}m{0.087\linewidth}>{\centering\arraybackslash\hspace{0pt}}m{0.087\linewidth}} 
\toprule
\rowcolor[rgb]{0.753,0.753,0.753} \multicolumn{1}{>{\centering\hspace{0pt}}m{0.15\linewidth}}{\textbf{Research Studies}} & \textbf{{\small Adversarial} Attack} & \textbf{Perturbation Generation} & \textbf{Universal Perturbation} & \textbf{Targeting Single Category} & \textbf{Interpretation-based Attack} & \textbf{Adaptive to Interpreters} & \textbf{White-box Attack} & \textbf{Black-box Attack} \\ 
\midrule
Moosavi \etal \cite{moosavi2017universal} & \checkmark & \checkmark & \checkmark &  &  &  & \checkmark &  \\ 

Khrulkov \etal \cite{khrulkov2018art} & \checkmark & \checkmark & \checkmark &  &  &  & \checkmark &  \\ 

Hayes \etal \cite{hayes2018learning} & \checkmark & \checkmark & \checkmark &  &  &  & \checkmark &  \\ 

Heo \etal \cite{heo2019fooling} & \checkmark &  &  &  & \checkmark &  & \checkmark &  \\ 

Zhang \etal \cite{zhang2020interpretable} & \checkmark & \checkmark &  &  & \checkmark & \checkmark & \checkmark &  \\ 

Eldor \etal \cite{10.1007/978-3-030-91434-9_9}  & \checkmark & \checkmark &  &  & \checkmark & \checkmark & \checkmark &  \\ 

Akhtar \etal \cite{akhtar2021explain}~~  & \checkmark & \checkmark & \checkmark & \checkmark &  &  & \checkmark &  \\

\textbf{Ours} & \textcolor{blue}{\checkmark} & \textcolor{blue}{\checkmark} & \textcolor{blue}{\checkmark} & \textcolor{blue}{\checkmark} & \textcolor{blue}{\checkmark} & \textcolor{blue}{\checkmark} & \textcolor{blue}{\checkmark} & \textcolor{blue}{\checkmark} \\
\bottomrule
\end{tabular}
}\vspace{3ex}
\end{table*}

\section{Related Work} \label{sec:related}

This section highlights previous studies related to our work, specifically in the domains of adversarial attacks and interpretation-guided attacks. 
Machine learning models face two primary threats: evasion attacks~\cite{he2020towards}, which involve manipulating the model's behavior through data manipulation, and poisoning attacks~\cite{he2020towards}, which weaken the target model by infecting the training data.
This work focuses on the first type of threat in which we manipulate data to make a DL model misbehave while ensuring the preservation of correct interpretation. Unlike existing approaches, we specifically consider targeted attacks where the perturbations affect a specific class without influencing other classes. Our work is one of the pioneering studies exploring targeted attacks against DL models using interpretability via a universal perturbation in both white-box and black-box settings. In the following, we highlight the related studies that are relevant to our work.

Moosavi-Dezfooli \etal \cite{moosavi2017universal} showed the existence of a universal and small-sized perturbation vector that can mislead the state-of-the-art deep neural networks. The work explored that there are single directions in the input space that can be used to generate universal noise. However, the universal perturbation can still be a noise-like pattern to the human eye when the interpretation is applied. The study \cite{akhtar2021explain} proposed that extended universal perturbation can exploit the explainability of models by carefully exploring the decision boundaries of deep models. The authors showed that their attack can be used to interpret the internal working process of DL models. Even though the attack is effective against the DL models, it is still vulnerable to interpreters. When an interpreter is adopted with a DL model, the involvement of the adversary can easily be detected (see \autoref{fig:overview}). The main idea of our attack is also based on generating universal perturbation \cite{akhtar2021explain}. Unlike those attacks, we consider misleading the interpretability along with classification while generating adversarial samples to increase the robustness and stealthiness of the attack.   

\BfPara{Interpretability} There have been numerous methods that have attempted to describe the inner working of deep learning models. Such methods operate by exploiting the characteristics of optimization methods (\eg back-propagation), intermediate representations, input manipulation and feature perturbation, and the development of meta models. 
Interpretability can offer a sense of security when inspecting the attribution maps using human involvement as adversarial examples reflect inaccurate attribution maps. However, recent studies show that some interpretation models are detached from DL models, and they can be impacted by manipulations without affecting the DL model performance \cite{kindermans2019reliability,ghorbani2019interpretation,heo2019fooling}.  

Other studies have demonstrated the validity and practicality of simultaneously attacking the deep learning models and their corresponding interpreters. 
These studies suggest that interpretability provides a limited sense of security \cite{zhang2020interpretable}. A recent work  \cite{10.1007/978-3-030-91434-9_9} introduces an optimized attack to fool IDLSes with limited perturbation using the edge information of the image. Similarly, our work focuses on attacking IDLSes; however, we consider generating universal perturbation instead of generating perturbation for each input.

The properties provided by \ours and previous studies are summarized in \autoref{tab:comparison}. 

\section{Methods} \label{sec:methods}
This section discusses several key concepts related to our work. We begin by presenting the problem formulation, which outlines the specific challenge we aim to address. We then describe the main algorithm for \ours.

\subsection{Fundamental Concepts} \label{sec:fund}
This section presents the concepts and notations, and key terms used throughout the paper. 

\BfPara{DL Model} As our paper mainly focuses on the classification task, let $f(x)=y \in Y$ denote a classifier that assigns an input $x$ to a category $y$ from a set of categories $Y$.

\BfPara{Interpreter} Let $g(x; f)= m$ denote an interpreter $g$ that generates an attribution map (\ie interpretation map) $m$ that reflects the importance of features in the input sample $x$ based on the output of the classifier $f$, (\ie the value of the $i\text{-th}$ element in $m$ reflects the importance of the $i\text{-th}$ element in $x$). Based on the methods used to obtain interpretations of a model, interpretations can be divided into two types: 

\begin{description}[leftmargin=1em]

\item \cib{1} \textbf{\ul{Pre-hoc Interpretability:}} \textcolor{black}{It is achieved by constructing self-explanatory models that integrate interpretability directly into their structures.} In other words, this type of interpretability focuses on building DL models that can explain their behavior explicitly in terms of the inference process \cite{zhang2018interpretable, sabour2017dynamic}. \textcolor{black}{The category includes decision tree, rule-based model, attention model, \etc}  

\item \cib{2} \textbf{\ul{Post-hoc Interpretability:}} Post-hoc Interpretability is based on the complexity-regulated DL model interpretation or adopting post-training methods \cite{murdoch2018beyond, dabkowski2017real}. This type of interpretation requires another model to provide explanations for the current model.
   
\end{description}

Our proposed attack specifically targets post-hoc interpretability, which involves the use of an interpreter that receives information from the target DL model (\eg gradients) and generates an interpretation of how the target model classifies an input sample. One reason for choosing this type of interpreter is that it does not require any modifications to the architecture of the prediction model, thereby preserving its high prediction accuracy.

{\BfPara{Threat Model}
This work considers both white-box and black-box attack scenarios. In the white-box attack setting, the adversary has complete access to the victim classifiers $f$ and their interpreters $g$. While white-box scenarios provide valuable insights into the strengths and weaknesses of the system, they are often impractical in real-world applications. Therefore, we also consider a black-box environment where the adversary has limited knowledge about the victim classifiers $f'$. 

\subsection{Target Interpretation Model}\label{sec:int_model}

The interpreters chosen for our experiment are representative of all state-of-the-art interpretation techniques. For example, Grad \cite{simonyan2014deep} shares the same formulations with DeepLift \cite{shrikumar2017learning}, SmoothGrad \cite{smilkov2017smoothgrad}, \etc, while CAM \cite{zhou2016learning} belongs to the same family of representation-guided interpreters (\eg GradCAM \cite{selvaraju2017grad}). 
We also use MASK interpreter \cite{fong2017interpretable} from the perturbation-guided interpreters.
Thus, the attack is applicable to other related interpreters.

\BfPara{CAM} In \textbf{C}lass \textbf{A}ctivation \textbf{M}ap (CAM) \cite{zhou2016learning}, interpretation maps are generated using feature maps taken from the target DL classifier's intermediate layers. The significance of the areas of the samples is generated by reflecting the weights of the fully-connected layer on the features maps of the convolutional layer.
Assume $a_{i}(j, k)$ as the activation of a channel $i$ in the last convolutional layer at a spatial position $(j, k)$, and $\sum_{j, k} a_{i}(j, k)$ denote the outcome of global average pooling.
So, the softmax function receives: 

\begin{equation*}
\begin{split}
    \psi_y(x) = \sum_{j, k}\sum_{i} w_{i, y} {}~ a_{i}(j, k),
\end{split}
\end{equation*}
and the attribution map $m_y$ is as follows:

\begin{equation*} 
\begin{split}
    m_{y}(j, k) = \sum_{i} w_{i, y} {}~ a_{i}(j, k),
\end{split}
\end{equation*}
where $w_{i, y}$ is the weight associated with the output class $y$ for $i$-th channel.
We construct interpretation maps by collecting and combining feature vectors from $f$ up to its last CNN layer.

\BfPara{Grad} To generate the importance of features of a given sample, the gradient of a DL classifier's outcome in terms of the given sample is computed by the interpreter. To be specific, based on the DL model $f$ and its prediction $y$ to a certain sample $x$, the interpreter generates an interpretation map or also called attribution map $m$ as $m_y = \left| \frac{\partial f_{y} (x)}{\partial x} \right|$.
Since the ReLU activation function is used in the target DL models, and all CNN-based models, the computed result of the Hessian matrix becomes all-zero. To find an optimal adversarial sample $\hat{x}$, the gradient of the ReLU $r(z)$ function is approximated as:

\begin{equation*} \label{eq:smoothRelu}
\begin{split}
    h(z) \triangleq \left\{ \begin{array}{lcr}
         (z+\sqrt{z^{2}+\tau})' = 1+\frac{z}{\sqrt{z^{2}+\tau}} & \mbox{for}
         & z < 0 \\
         (\sqrt{z^{2} + \tau})' = \frac{z}{\sqrt{z^{2} + \tau}} & \mbox{for} & z \geq 0
                \end{array}\right. 
\end{split}
\end{equation*}
where $h(z)$ approximates the gradient of ReLU $r(z)$ with a small constant parameter $\tau$ (\eg $\tau = 1e-4$ ) \cite{zhang2020interpretable}. 

\BfPara{MASK} The interpreter \cite{fong2017interpretable} generates interpretation maps by detecting changes in the prediction of a DL model while adding a minimal amount of noise to the sample. Specifically, the interpreter creates a mask $m$ that is a binary matrix of the same size as the sample $x$. In the matrix, 0 represents the area of the sample $x$ where the feature is kept without noise. The value 1 in the matrix means that the area is replaced with Gaussian noise. The main objective of the interpreter is to find the smallest mask $mask$ that makes a DL model's performance decrease greatly:

\begin{equation} \label{eq:maskFormula1}
\begin{split}
     \min_{\text{\tiny\em mask}} : f_{y}(\phi(x; \text{\tiny\em mask})) + \lambda ~\Vert 1 - \text{\scriptsize\em mask} \Vert _{1} \quad s.t. \quad 0 \leq \text{\tiny\em mask} \leq 1,
\end{split}
\end{equation}
where $\phi(x; \text{\scriptsize\em mask})$ is the operator that generates perturbation to decrease the probability of the current prediction category $y$ and the second term $\lambda ~\Vert 1 - \text{\scriptsize\em mask} \Vert$ helps the mask to be scattered.
Since the MASK interpreter is an optimization function, applying the attack (as another optimization) is directly infeasible. We reformulate the attack as a bi-level optimization task (similar to the framework in \cite{zhang2020interpretable, abdukhamidov2021advedge,abdukhamidov2022black}). 

\subsection{Problem Formulation} \label{sec:problem}
Let $S$ be a distribution over the dataset and $\mathbf{s}$ $\in$ $\mathbb{R}^d$ indicates a sample from a distribution $S$. For trained model $f(s) \rightarrow y$, where $y$ is the correct class, 
the main objective of adversarial attacks is to generate a perturbation ${p}$ $\in$ $\mathbb{R}^d$ that satisfies the following constraint:
\begin{equation} \label{eq:advFormula}
\begin{aligned}
    f(\mathbf{s} + {p}) \rightarrow y_t, where \: y \neq y_t, \: \|{p}\|_{\ell_p} \leq \eta
\end{aligned}
\end{equation}

In \autoref{eq:advFormula}, confining $y_t$ to a chosen category and $\ell_p$-norm vector to a pre-set value of $\eta$ produces a targeted attack.

Generating universal perturbation in adversarial attacks expands the domain of ${p}$, which we denote as $D({p})$. Given that $|D({p})| \geq 1$, where ~$|~.~|$ denotes the cardinality of a set, we maximize the objective of \autoref{eq:advFormula} as follows:
\begin{equation} \label{eq:maxFormula}
\begin{split}
    \max ~\textbf{P}_{D({p})}(f(\mathbf{s} + {p}) \rightarrow y_t) \geq \gamma, \\ where \: \|{p}\|_{\ell_p} \leq \eta, and \: |D(\mathbf{p})| \geq 1 
\end{split}
\end{equation}

In \autoref{eq:maxFormula}, $\textbf{P}$ refers to the probability distribution of generating $p$, and $\gamma$ is the attack success threshold with a fixed value in the range of $[0,1]$.

In \ours{} case, we consider an interpretation model to generate universal perturbation that keeps the adversarial interpretation result similar to the interpretation of the original samples. Hence, we have the following constraints:

\begin{description}[leftmargin=1ex,itemsep=1ex]
    \item \cib{1} Ensuring model misclassification to a pre-defined category: $f(\mathbf{s} + {p})$ $\rightarrow$ $y_t$, $y$ $\neq$ $y_t$, where $y_t$ is the target category.
    
    \item \cib{2} Restricting the sample domain to the selected category: $D({p})= \{\mathbf{s}~|~\mathbf{s} ~\sim S_{\text{\tiny \em selected}}\}$, where $S_{\text{\tiny \em selected}}$ the distribution of the targeted category. 
    
    \item \cib{3} Restricting the effect of the perturbation on other categories' domain: $\textbf{P}_{\hat{D}({p})}(f(\mathbf{s} + {p}) \rightarrow y_t) < \gamma$, where $\hat{D}(p)$ denotes the domain of samples from non-selected categories.
    
    \item \cib{4} Triggering an interpreter $g$ to generate target attribution maps: $g(\mathbf{s} + {p}; f) ~~\xrightarrow{\textit{\small{similar}}}
 m_t$, such that $g(\mathbf{s}; f) \rightarrow m_t$.
\end{description}

The following subsection describes the attack algorithm and the considerations to meet the outlined constraints. 


\subsection{Computing the Perturbation}\label{sec:com_pert}

\autoref{alg:algorithm_pertb} describes the perturbation generation and follows the constraints mentioned in \autoref{sec:problem}. As the objective of the algorithm is straightforward, a universal perturbation is calculated by steps taken to minimize the cost of the attack such that the used classifier and the interpreter increase the confidence of adversarial samples for the selected category while minimizing the difference between benign and adversarial attribution maps. 

The desired cost surface for high confidence and low interpretation loss is based on stochastic computation and is ruled by the first and second-moment estimations. The first and the second moment check if the computed perturbation prevents other categories from crossing their decision boundaries during the generation process. The computed perturbation should not interfere with the prediction of non-source classes. $\ell_{\infty}$ norm is applied to bounce the perturbation norm. We explain each line of the algorithm in detail.

The \textit{typical} attack is based on the white-box scenario as it requires the target model's parameters. Samples of $X_{\text{\tiny \em selected}}$ and $\hat{X}$ are accumulated from $D({p})$ and $\hat{D}({p})$, \ie samples from the selected categories and other categories,  respectively. 
Like other parameters, $\eta$ for $\ell_p$-norm of the perturbation, target category $y_t$, target attribution map $m_t$, batch size $b$ for the optimization, fooling ratio $\gamma$ (confidence level as a target category), pre-set first, and second-moment hyper-parameters are considered (\textit{line 1}).

In the algorithm, firstly, the selected and the other categories are sampled randomly into sets $X_x$ and $X_o$, and the cardinality of each set equals half of the batch size $b$ (\textit{line 3}). On \textit{line 4}, all sets are perturbed by subtracting the currently estimated perturbation $p_t$ from each of them (the operation is displayed as $\ominus$). Afterward, the perturbed sets are clipped to a valid range by the $C$ function.

\RestyleAlgo{ruled}
\SetKwComment{Comment}{/* }{ */}
\newcommand\mysim{\mathrel{\stackrel{\makebox[0pt]{\mbox{\normalfont\tiny rand}}}{\sim}}}

\begin{algorithm}[t]
\small
\caption{\ours{} attack's main algorithm}\label{alg:algorithm_pertb}

\KwData{Target model $f$, interpreter $g$, selected category samples $X_{\text{\tiny \em selected}}$, non-selected categories samples $\hat{X}$ $s.t. ~ \hat{X}_i ~\text{or}~ X_i \in \mathbb{R}^d$, target category $y_{t}$, target attribution map $m_t$, perturbation norm $\eta$, balance factor $\lambda$, batch size $b$, and fooling ratio $\gamma$,  $\beta_1 = 0.9$ and $\beta_2 = 0.999$.} 

\KwResult{Perturbation $p$ $\in$ $\mathbb{R}^d$}
\textbf{Initialization:} Setting $p_0, \upsilon_0 \in \mathbb{R}^d, \omega_0 \in \mathbb{R}^d$ and $i=0$\;

 \While{fooling ratio $<$ $\gamma$}
 {
  $X_x$ $\mysim$ $X$, $X_o$ $\mysim$ $\hat{X}$ : $|X_x|$ = $|X_o|$ = $\frac{b}{2}$\;
  $S_x$ $\gets$  C($X_x$ $\ominus$ $p_i$), $S_o$ $\gets$  C($X_o$ $\ominus$ $p_i$)\;
  $i \gets i + 1$\;
  
  $\delta \gets \frac{\mathbb{E}_{x_i \in S_x} \big[\| \nabla_{x_i} (L_{prd}(f(x_i, y_{t})) +  \lambda L_{int}(g(x_i; f), m_t))\|_2\big]}
                    {\mathbb{E}_{x_i \in S_o} \big[\| \nabla_{x_i} (L_{prd}(f(x_i, y)) + \lambda L_{int}(g(x_i; f), m_t))\|_2\big]}$\;\vspace{1ex}

  $\xi_i \gets \frac{1}{2} \textbf{\Big(}\mathbb{E}_{x_i \in S_x} \big[ \nabla_{x_i} (L_{prd}(f(x_i, y_{t})) + 
  \newline \hspace*{8em} \lambda L_{int}(g(x_i; f), m_t))\big]+ 
  \newline \hspace*{3.3em} \delta \mathbb{E}_{x_i \in S_o} \big[ \nabla_{x_i}( L_{prd}(f(x_i, y)) + 
  \newline \hspace*{8em} \lambda L_{int}(g(x_i; f), m_t))\big]\textbf{\Big)}$\;\vspace{1ex}
  
  $\upsilon_i \gets \beta_1 \upsilon_{i-1} + (1 - \beta_1) ~\xi_i$\;
  $\omega_i \gets \beta_2 \omega_{i - 1} + (1 - \beta_2)(\xi_i \odot \xi_i)$\;\vspace{1ex}
  $\bar{p} \gets \frac{\sqrt{1 - \beta^{i}_2}}{1-\beta^{i}_1} .~ diag\big(diag(\sqrt{\omega_i})^{-1} \upsilon_i\big)$\;
  $p_i \gets p_{i - 1} + \frac{\bar{p}}{\|\bar{p}\|_\infty}$\;
  $p_i \gets sign(p_i) \odot \min (|p_i|, \eta)$\;\vspace{1ex}
 }
\end{algorithm}

On \textit{line 6}, we calculate the ratio between the expected norms \textcolor{black}{ (it is referred to as $\mathbb{E}$ in the algorithm)} of the selected category gradients and other categories' gradients by calculating the expected gradient of an input using the selected and non-selected category samples, and the computed attribution maps of the interpreter. 
In the algorithm, $L_{prd}$ is the classification loss function that shows the difference between the model prediction and the target category. $L_{int}$ is the interpretation loss for calculating the difference between the current and target attribution maps: $L_{int}(g(x; f), m_t)) = \|(g(x; f) - m_t\|_{2}^{2}$. $\lambda$ balances the two factors. The value of the hyper-parameter depends on the interpreter. In our experiments, we explored different values of $\lambda$ to account for different interpreters. 

With those functions, we calculate the $\xi_i$ as the average of the expected gradients for both source and non-source classes (\textit{line 7}), which keeps the direction of the vector to obtain the targeted fooling for the source classes while preventing the fooling of non-source classes. At the same time, it considers the interpretation loss in choosing the optimal vector direction. Specifically, $L_{int}$ evaluates the difference between the current and the desired interpretation maps based on the chosen vector direction. This, in turn, helps achieve the desired result formulated in \autoref{sec:problem}. Next, on \textit{lines 8 and 9}, the first and the second raw moment (\ie un-centered variance) of the computed gradient \textcolor{black}{(referred to as $\upsilon$ and $\omega$ respectively)} are calculated with moving averages exponentially ($\odot$ represents the Hadamard product) for the effective stochastic optimization on the cost surface. 

On \textit{line 10}, we conduct bias-corrected estimation since the second moment (\ie moving average) is known to be heavily biased in the early stages of the optimization. The derivation of the expression in \textit{line 10 }is explained in \cite{akhtar2021explain}. The perturbation is computed by the ratio between the moment estimates ($\frac{\upsilon_i}{\sqrt{\omega_i}}$, where ${\omega_i}$ represents the second moment), and as the notations are vectors, we convert vectors into diagonal matrices or diagonal matrices into vectors via $diag(.)$ operation. 

Finally, we update the perturbation by restricting the $\bar{p}$ with $\ell_\infty$-norm to keep the desired direction (\textit{line 11}). The norm of the computed perturbation is restricted by $\ell_\infty$-ball projection at the end of each iteration \textcolor{black}{to minimize the perturbation perceptibility by performing the Hadamard ($\odot$) product between the element-wise sign  ($sign(.)$) and the minimum values for perturbation (\textit{line 12}).}

The objective of \ours{} is not only to deceive the classifier by making it misclassify the designated source class samples while correctly predicting non-source classes but also to preserve the original interpretation for all samples of the source class. From an adversarial standpoint, this approach minimizes suspicion of the attack by manipulating a single class rather than making the classifier predict the same label for all images and providing different attribution maps.


\section{\ours{} in White-box Settings} \label{sec:evaluation}

This section presents the experimental results obtained in white-box settings. We evaluate the effectiveness of the generated perturbation from two perspectives: adversarial success and interpretation validity.

\subsection{Experimental Settings} \label{sec:settings}

\BfPara{Dataset} We use the ImageNet dataset \cite{ILSVRC15} for our experiments. The training set consists of 1,300 samples per category from the ImageNet dataset. We use the training set within the attack framework to calculate the perturbation. The testing portion of the dataset, which includes 50 samples per category (both source and non-source categories), is used to evaluate the generated perturbation. 
To ensure accurate gradient directions, we randomly select samples that are correctly classified with $\geq$60\% confidence from both the targeted and non-targeted categories. In this paper, we designate certain object classes as the targeted category while considering other classes as the non-targeted category. For instance, categories such as \textbf{panda}, \textbf{dog}, and \textbf{cup} are used as targeted categories (see \autoref{tab:fooling_ratio}).

\BfPara{DL Models} In the experiments, four state-of-the-art DL models are used for our targeted attack, which are \textbf{ResNet-50} \cite{he2016deep} \textcolor{black}{(74.90\% top-1 accuracy and 92.10\% top-5 accuracy on ImageNet dataset), \textbf{VGG-16} \cite{simonyan2014very} (71.30\% top-1 accuracy and 90.10\% top-5 accuracy on ImageNet dataset)}, \textbf{DenseNet-169} \cite{huang2017densely} (76.20\% top-1 accuracy and 93.15\% top-5 accuracy on ImageNet dataset), and \textbf{Inception-V3} \cite{szegedy2016rethinking} (77.90 \% top-1 accuracy and 93.70\% top-5 accuracy on ImageNet dataset). The models are chosen in terms of performance and network architecture to help measure the effectiveness of our attack. 

\BfPara{Interpretation Models} \textbf{CAM} \cite{zhou2016learning}, \textbf{Grad} \cite{simonyan2014deep} and \textbf{MASK} \cite{fong2017interpretable} interpreters are utilized as the representative of the interpretation models. Their original open-source implementations are used for our experiment.

\begin{table*}
\centering
\caption{Fooling ratio, misclassification confidence, and leakage rate against several models using ImageNet. 
The results show source category $\xrightarrow[]{\text{transformed}}$ target category for Target 1: Panda $\rightarrow$ \textbf{Cat}, Target 2: Dog $\rightarrow$ \textbf{Goose}, Target 3: Cup $\rightarrow$ \textbf{Wolf}.}
\label{tab:fooling_ratio}

\arrayrulecolor{black}
\resizebox{\linewidth}{!}{%
\begin{tabular}{ccccc|ccc|ccc|ccc} 
\toprule
\rowcolor[rgb]{0.753,0.753,0.753} {\cellcolor[rgb]{0.753,0.753,0.753}}                                       & {\cellcolor[rgb]{0.753,0.753,0.753}}                                 & \multicolumn{3}{c|}{\textbf{Target 1}}                                                                                                                                                                                                                                                                                                                         & \multicolumn{3}{c|}{\textbf{Target 2}}                                                                                                                                                                                                                                                                                                                                          & \multicolumn{3}{c|}{\textbf{Target 3}}                                                                                                                                                                                                                                                                                                                                          & \multicolumn{3}{c}{\textbf{Average}}                                                                                                                                                                                                                                                                                                                                            \\ 

\rowcolor[rgb]{0.753,0.753,0.753} \multirow{-2}{*}{{\cellcolor[rgb]{0.753,0.753,0.753}}\textbf{Interpreter}} & \multirow{-2}{*}{{\cellcolor[rgb]{0.753,0.753,0.753}}\textbf{Model}} & \begin{tabular}[c]{@{}>{\cellcolor[rgb]{0.753,0.753,0.753}}c@{}}\textbf{Fooling }\\\textbf{Ratio}\end{tabular} & 
\begin{tabular}[c]{@{}>{\cellcolor[rgb]{0.753,0.753,0.753}}c@{}}\textbf{Misclassification }\\\textbf{Confidence}\end{tabular} & \begin{tabular}[c]{@{}>{\cellcolor[rgb]{0.753,0.753,0.753}}c@{}}\textbf{Leakage }\\\textbf{Rate}\end{tabular} & \begin{tabular}[c]{@{}>{\cellcolor[rgb]{0.753,0.753,0.753}}c@{}}\textbf{Fooling }\\\textbf{Ratio}\end{tabular} & \begin{tabular}[c]{@{}>{\cellcolor[rgb]{0.753,0.753,0.753}}c@{}}\textbf{Misclassification }\\\textbf{Confidence}\end{tabular} & \begin{tabular}[c]{@{}>{\cellcolor[rgb]{0.753,0.753,0.753}}c@{}}\textbf{\textbf{Leakage}}\\\textbf{\textbf{Rate}}\end{tabular} & \begin{tabular}[c]{@{}>{\cellcolor[rgb]{0.753,0.753,0.753}}c@{}}\textbf{Fooling }\\\textbf{Ratio}\end{tabular} & \begin{tabular}[c]{@{}>{\cellcolor[rgb]{0.753,0.753,0.753}}c@{}}\textbf{Misclassification }\\\textbf{Confidence}\end{tabular} & \begin{tabular}[c]{@{}>{\cellcolor[rgb]{0.753,0.753,0.753}}c@{}}\textbf{\textbf{Leakage}}\\\textbf{\textbf{Rate}}\end{tabular} & \begin{tabular}[c]{@{}>{\cellcolor[rgb]{0.753,0.753,0.753}}c@{}}\textbf{Fooling }\\\textbf{Ratio}\end{tabular} & \begin{tabular}[c]{@{}>{\cellcolor[rgb]{0.753,0.753,0.753}}c@{}}\textbf{Misclassification }\\\textbf{Confidence}\end{tabular} & \begin{tabular}[c]{@{}>{\cellcolor[rgb]{0.753,0.753,0.753}}c@{}}\textbf{\textbf{Leakage}}\\\textbf{\textbf{Rate}}\end{tabular}  \\ 
\midrule
\arrayrulecolor{black}
\multirow{4}{*}{\textbf{CAM}}                                                                                & VGG-16                                                               & 0.85                                                                                                           & 0.82                                                                                                                          & 0.30                                                                                                          & 0.87                                                                                                           & 0.85                                                                                                                          & 0.35                                                                                                                           & 0.71                                                                                                           & 0.80                                                                                                                          & 0.32                                                                                                                           & 0.81 ± 0.07                                                                                                    & 0.82                                                                                                                          & 0.32                                                                                                                            \\ 
                                                                                                             & ResNet-50                                                            & 0.86                                                                                                           & 0.86                                                                                                                          & 0.36                                                                                                          & 0.79                                                                                                           & 0.82                                                                                                                          & 0.34                                                                                                                           & 0.75                                                                                                           & 0.79                                                                                                                          & 0.35                                                                                                                           & 0.80 ± 0.05                                                                                                    & 0.82                                                                                                                          & 0.35                                                                                                                            \\ 
                                                                                                             & DenseNet-169                                                         & 0.78                                                                                                           & 0.81                                                                                                                          & 0.34                                                                                                          & 0.75                                                                                                           & 0.80                                                                                                                          & 0.33                                                                                                                           & 0.69                                                                                                           & 0.77                                                                                                                          & 0.36                                                                                                                           & 0.74~± 0.05                                                                                                    & 0.79                                                                                                                          & 0.34                                                                                                                            \\ 
                                                                                                             & Inception-V3                                                         & 0.68                                                                                                           & 0.79                                                                                                                          & 0.30                                                                                                          & 0.62                                                                                                           & 0.78                                                                                                                          & 0.31                                                                                                                           & 0.60                                                                                                           & 0.75                                                                                                                          & 0.29                                                                                                                           & 0.63~± 0.04                                                                                                    & 0.77                                                                                                                          & 0.30                                                                                                                            \\ 
\midrule
\multirow{4}{*}{\textbf{\textbf{Grad}}}                                                                      & VGG-16                                                               & 0.87                                                                                                           & 0.71                                                                                                                          & 0.34                                                                                                          & 0.85                                                                                                           & 0.75                                                                                                                          & 0.36                                                                                                                           & 0.75                                                                                                           & 0.77                                                                                                                          & 0.31                                                                                                                           & 0.82 ± 0.03                                                                                                    & 0.74                                                                                                                          & 0.34                                                                                                                            \\ 
                                                                                                             & ResNet-50                                                            & 0.83                                                                                                           & 0.88                                                                                                                          & 0.38                                                                                                          & 0.81                                                                                                           & 0.85                                                                                                                          & 0.37                                                                                                                           & 0.72                                                                                                           & 0.76                                                                                                                          & 0.35                                                                                                                           & 0.79 ± 0.02                                                                                                    & 0.83                                                                                                                          & 0.37                                                                                                                            \\ 
                                                                                                             & DenseNet-169                                                         & 0.80                                                                                                           & 0.86                                                                                                                          & 0.35                                                                                                          & 0.73                                                                                                           & 0.82                                                                                                                          & 0.33                                                                                                                           & 0.70                                                                                                           & 0.76                                                                                                                          & 0.35                                                                                                                           & 0.74~± 0.02                                                                                                    & 0.81                                                                                                                          & 0.34                                                                                                                            \\ 
                                                                                                             & Inception-V3                                                         & 0.72                                                                                                           & 0.80                                                                                                                          & 0.31                                                                                                          & 0.68                                                                                                           & 0.79                                                                                                                          & 0.33                                                                                                                           & 0.65                                                                                                           & 0.72                                                                                                                          & 0.30                                                                                                                           & 0.68 ± 0.01                                                                                                    & 0.77                                                                                                                          & 0.31                                                                                                                            \\ 
\midrule
\multirow{4}{*}{\textbf{\textbf{MASK}}}                                                                      & VGG-16                                                               & 0.81                                                                                                           & 0.79                                                                                                                          & 0.36                                                                                                          & 0.76                                                                                                           & 0.77                                                                                                                          & 0.34                                                                                                                           & 0.72                                                                                                           & 0.70                                                                                                                          & 0.33                                                                                                                           & 0.76 ± 0.04                                                                                                    & 0.75                                                                                                                          & 0.34                                                                                                                            \\ 
                                                                                                             & ResNet-50                                                            & 0.80                                                                                                           & 0.82                                                                                                                          & 0.37                                                                                                          & 0.73                                                                                                           & 0.84                                                                                                                          & 0.36                                                                                                                           & 0.70                                                                                                           & 0.72                                                                                                                          & 0.34                                                                                                                           & 0.74 ± 0.04                                                                                                    & 0.79                                                                                                                          & 0.36                                                                                                                            \\ 
                                                                                                             & DenseNet-169                                                         & 0.77                                                                                                           & 0.81                                                                                                                          & 0.35                                                                                                          & 0.71                                                                                                           & 0.83                                                                                                                          & 0.33                                                                                                                           & 0.68                                                                                                           & 0.75                                                                                                                          & 0.34                                                                                                                           & 0.72~± 0.04                                                                                                    & 0.80                                                                                                                          & 0.34                                                                                                                            \\ 
                                                                                                             & Inception-V3                                                         & 0.69                                                                                                           & 0.76                                                                                                                          & 0.30                                                                                                          & 0.64                                                                                                           & 0.79                                                                                                                          & 0.34                                                                                                                           & 0.63                                                                                                           & 0.70                                                                                                                          & 0.31                                                                                                                           & 0.65 ± 0.03                                                                                                    & 0.75                                                                                                                          & 0.32                                                                                                                            \\
\bottomrule
\end{tabular}
}
\end{table*}


\BfPara{Attack Evaluation} 
The effectiveness of the attack is evaluated by conducting several experiments and calculating some evaluation metrics in terms of attack success rate in fooling the classifier while maintaining a convincing interpretation.
The evaluation aims to find answers to the following questions:
\cib{1} \textit{Is our technique effective in attacking DL models by targeting a single category? }
\cib{2}\textit{Is our technique effectively misleading the interpreters by generating a similar interpretation to the benign sample?}

\BfPara{Evaluation Metrics} To assess the effectiveness of the proposed attack against the target IDLSes, we utilize various metrics. These metrics are employed to evaluate the attack's impact on both the classifiers and interpreters. The following metrics are used in our evaluation: 

\noindent\textbf{Against Classifiers:}
\begin{itemize}[itemsep=0.5ex,leftmargin=1em]
    \item \textbf{Fooling Ratio:} This metric \cite{moosavi2017universal} measures the proportion of images that undergo the universal perturbation and have their labels changed. It indicates the attack's success in causing misclassifications by the target model, specifically into the target category. This metric provides a quantitative measure of the attack success against DL models.
    
    \item \textbf{Misclassification Confidence:} This metric \cite{zhang2020interpretable} measures the probability (\ie \textit{confidence score}) of an adversarial sample assigned by the target model to the target category (\ie we calculate the average confidence scores of adversarial samples successfully misclassified).
    
    \item \textbf{Classification Confidence:} This metric \cite{zhang2020interpretable} evaluates the impact of universal perturbations on non-target categories. It measures the confidence scores of these categories when the perturbation is applied, revealing the extent to which the model's performance is affected while maintaining its effectiveness against the targeted category. 
    
    \item \textbf{Leakage Rate:} This metric \cite{moosavi2017universal} measures the impact of universal perturbations on non-source categories. It is calculated as the ratio of misclassified non-source category images to the total number of non-source category images used for testing when the universal perturbation is applied. A lower leakage rate indicates that the universal perturbations are more specific to the targeted category, while a higher leakage rate suggests a more general impact.
\end{itemize}

\noindent\textbf{Against Interpreters:}

\begin{itemize}[itemsep=0.5ex,leftmargin=1em]

    \item \textbf{Qualitative Comparison:} This method \cite{zhang2020interpretable} is used to verify whether the results of interpretation are perceptually similar. Every interpretation map is manually checked to see if it is similar to its benign interpretation map or if the interpretation is reliable.
    \item \textbf{IoU Test} (\textbf{I}ntersection-\textbf{o}ver-\textbf{U}nion): This metric \cite{he2017mask} is used to quantify the similarity between two arbitrary shapes. It encodes the shape properties of interpretation maps, \eg \textit{height, width, and location} into region properties and calculates the intersection areas between the predictions and the ground truths. It is widely employed to evaluate object detection, segmentation, and tracking: 
    $$\text{IoU}(m, m_{\circ}) = | O(m) \bigcap O(m_{\circ})| ~ / ~| O(m) \bigcup  O(m_{\circ}) |.$$ 
    
    In the formula, $m$ represents the attribution map of samples when the universal perturbation is added and $m_{\circ}$ is the attribution map of samples without any perturbation, and $O(.)$ represents a binarization function. In our case, we compare an adversarial interpretation map with a benign interpretation map based on (shapes, positions, and areas), to which the metric can be applied.

\end{itemize}

\begin{figure}
    \centering
    \captionsetup{justification=justified}
    \includegraphics[width=1\linewidth]{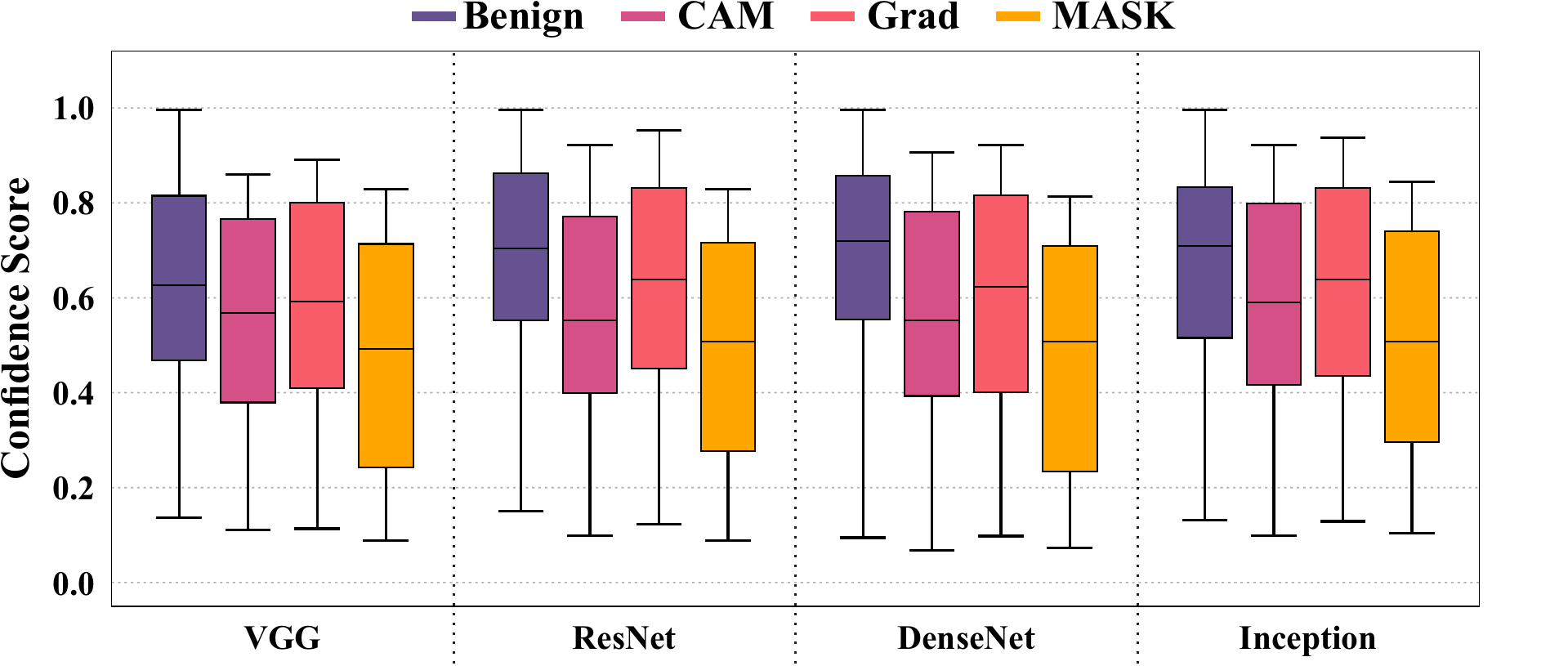}
    \caption{Classification confidence of adversarial samples (non-source categories) based on VGG-16, ResNet-50, DenseNet-169, and Inception-V3 with CAM, Grad, and MASK interpreters. The universal perturbation is for {Dog} $\rightarrow$ \textbf{Goose}.}
    \label{fig:confidence_score}
\end{figure}

\begin{figure*}[t]
    \centering
    \captionsetup{justification=justified}
    \includegraphics[width=0.75\linewidth]{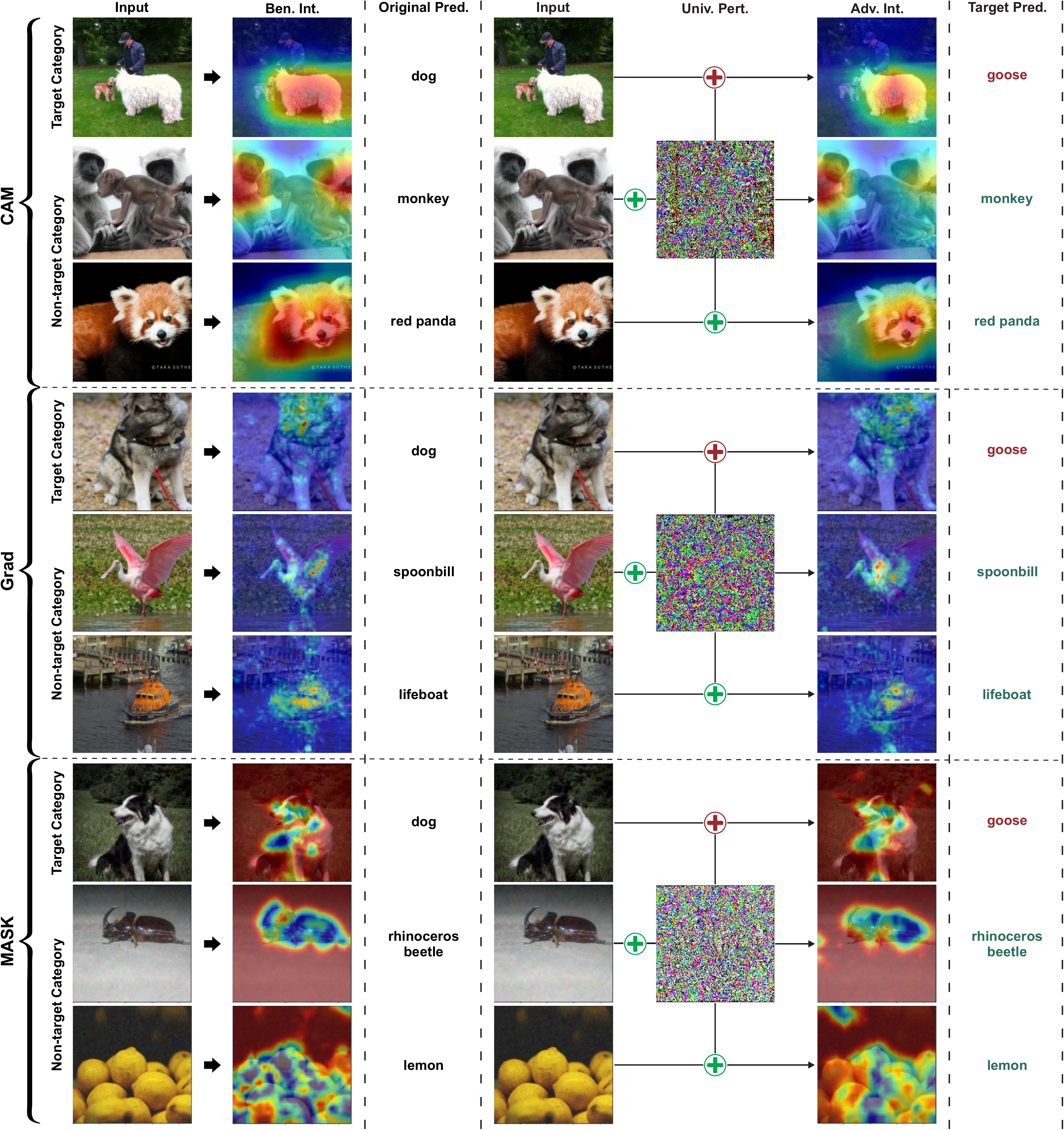}
    \caption{Attribution maps of benign and adversarial samples based on VGG-16, ResNet-50 with CAM, Grad and MASK interpreters. In this sample, our target category is Dog $\rightarrow$ \textbf{Goose}. (Ben.Int. stands for Benign Interpretation, Adv.Int. stands for Adversarial Interpretation and Univ.Pert. stands for Universal Perturbation).}
    \label{fig:int_samples}
    \vspace{-2ex}
\end{figure*}

\subsection{Attack Effectiveness Against DL Models}
We first assess the effectiveness of the attack in terms of deceiving the classifiers. The results are summarized using the \textit{fooling ratio}, \textit{misclassification confidence}, and \textit{leakage rate} in \autoref{tab:fooling_ratio}.
The reported results (\ie fooling ratio or attack success rate, misclassification confidence, and leakage rate) are based on the test samples that are not seen by the selected model and the attack algorithm. The success of the attack is demonstrated by fooling ResNet-50, VGG-16, DenseNet-169, and Inception-V3 trained on the ImageNet dataset \cite{ILSVRC15}. 

Observing the results for four classifiers (\ie ResNet-50, VGG-16, DenseNet-169, and Inception-V3), the attack generates a universal perturbation for different architectures with high fooling rates. More specifically, VGG-16 and ResNet-50 were deceived with the universal perturbation of \ours{} with a success rate of more than 70\% for all target categories regardless of the interpreter. This means that the addition of the universal perturbation to any raw image in our test samples can deceive the target DL models more than seven times out of ten. The attack also achieved significantly better results with a higher than 60\% fooling ratio in all interpreters when Densenet-169 and Inception-V3 were used as the target DL models. Among the target models, Inception-V3 was attacked with a relatively lower fooling ratio while VGG-16 achieved a higher fooling ratio in comparison with the other models.

Based on the \textit{misclassification confidence}, the attack fooled target DL models with confidence scores higher than 70\% regardless of the interpreters employed. Upper results can be seen when the attack is implemented against ResNet-50 and DenseNet-169, while the Inception-V3 model provides lower results across all interpreters. The main reason for the lower misclassification confidence scores of Inception-V3 could be due to the global averaging before the output layer, which reduces the computational cost and diminishes the effect of the perturbation on the output. 

According to the analysis of the leakage rate, the proposed adversarial algorithm appears to perform well. The computed leakage rate indicates that the algorithm's universal perturbation has a limited impact on non-source classes, with an average of 33\% leakage rate across all interpreters while the existing attack \cite{akhtar2021explain} has an average leakage rate of 41.5\%. This suggests that the perturbation is more specific to the targeted source class and has minimal impact on other non-source classes, which is a desirable characteristic. A low leakage rate implies that the algorithm's universal perturbation has a more targeted effect and is less likely to cause errors in the classification of non-source classes.

The results of the classification confidence metric are presented in \autoref{fig:confidence_score}, which depicts the impact of a universal perturbation on the confidence scores assigned by the DL models for non-source categories when various interpreters are used. To provide a basis for comparison, the scores obtained without any universal perturbation are also included. The figure shows that the universal perturbations generated using the CAM and Grad interpreters have a smaller impact on non-source categories than that of the MASK interpreter. Nevertheless, the results indicate that the perturbation has a minimal impact on the scores of non-target categories.

Observing the performance of all classifiers under attack, \ours{} generates a universal perturbation for different architectures with high fooling rates. These results suggest that \ours{} has successfully fooled the target DL models considering a single-class targeted attack scenario. 

\subsection{Attack Effectiveness Against the Interpreters}
In this section, we evaluate the effectiveness of the attack to generate adversarial samples that produce attribution maps similar to the benign samples using the targeted interpreter. 
Firstly, we use qualitative comparison to verify if the produced attribution maps of adversarial samples are perceptually similar to their benign samples. We checked all the adversarial attribution maps and found that observing all the cases for all targeted categories, \ours{} attack generates universal perturbations that produce attribution maps on adversarial domains similar to or indistinguishable from the attribution maps of the corresponding benign domain. 

\autoref{fig:int_samples} displays a set of samples alongside their attribution maps based on the \textbf{CAM}, \textbf{Grad}, and \textbf{MASK} interpreters. The samples are selected randomly from the output set. In the figure, the first three columns display the benign samples, their attribution maps, and their prediction categories. The last three columns present universal perturbations generated based on a DL model and an adopted interpreter, adversarial attribution maps produced by adding the universal perturbations to the benign samples, and target prediction categories. As shown in the figure, the results support high similarity in terms of interpretations. By observing the produced attribution maps for adversarial samples in both targeted and non-targeted categories, \ours{} produces perturbations that only affect the target category in terms of prediction while maintaining accurate interpretations across all categories. 

\begin{figure}[t]
    \centering
    \captionsetup{justification=justified}
    \includegraphics[width=\linewidth]{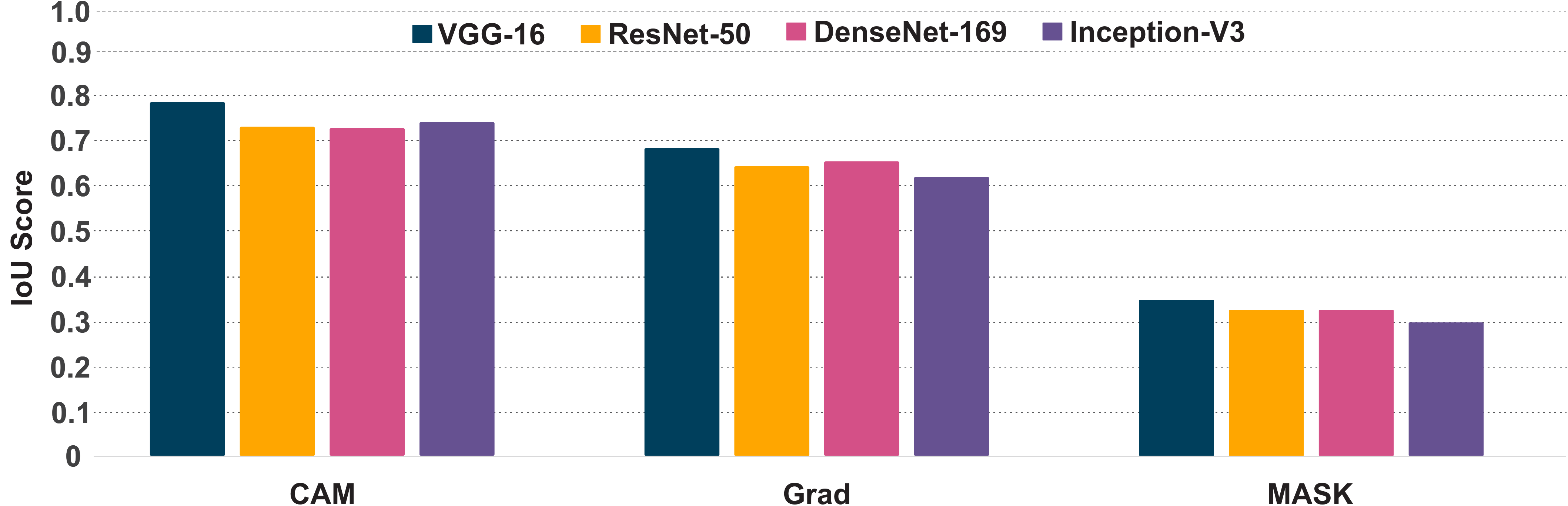}
    \caption{IoU scores for attribution maps of adversarial inputs using different thresholds. The results are based on CAM, Grad, and MASK interpreters for VGG-16, ResNet-50, DenseNet-169, and VGG-16.}
    \label{fig:IOU_sample}
\end{figure}

Additionally, the attribution maps of benign samples and adversarial samples (when the universal perturbation is added to the benign images) are compared using the IoU score metric. 
The IoU score is a quantitative measure that compares model outputs to ground-truth data. In our case, model outputs are generated attribution maps of adversarial samples, while ground-truth data contains benign attribution maps. To estimate the IoU score, the attribution maps that contain floating numbers in the range of [0, 1] are binarized, \ie their fundamental values are assigned one or zero based on a threshold value. All values higher than a threshold are assigned to 1; otherwise, the value is set to 0. We applied different threshold values (\textit{0.1, 0.2, 0.3, 0.4, 0.5, 0.6, 0.7, 0.8, 0.9}) to measure the IoU scores with attribution maps and calculated the average values of those IoU scores. Region of interest (RoI) is generally considered positive if it has an IoU score of 0.5 or above compared with the ground-truth sample. Thus, we can assume that adversarial interpretation maps with an IoU score equal to or above 0.5 based on their benign interpretation maps are credible. However, we also check if interpretation maps with low IoU scores are meaningful regarding their images.

The IoU scores of attribution maps from different DL models are displayed in \autoref{fig:IOU_sample}. 
Using CAM and Grad interpreters, the achieved IoU scores of all models (\ie VGG-16, ResNet-50, DenseNet-169, and Inception-V3) are above 60\% across all thresholds when the universal perturbation is added to the samples on the test set. Using the MASK interpreter, the attack achieved significantly lower IoU scores across all DL models. The main reason for the lower IoU scores in the MASK is that benign and adversarial attribution maps do not share high similarities regarding the shape and size of highlighted areas. However, based on our inspection of many generated adversarial examples, we observed that the adversarial attribution maps are similar to the benign ones with respect to their highlighted region and position in sample images. \autoref{sec:limiation} discusses other factors for the lower IoU scores in detail. The results show that the adversarial examples generate highly similar or meaningful attribution maps to the corresponding benign samples even when restricting the attribution maps to higher values.

\begin{figure}
    \centering
    \captionsetup{justification=justified}
    \includegraphics[width=0.9\linewidth]{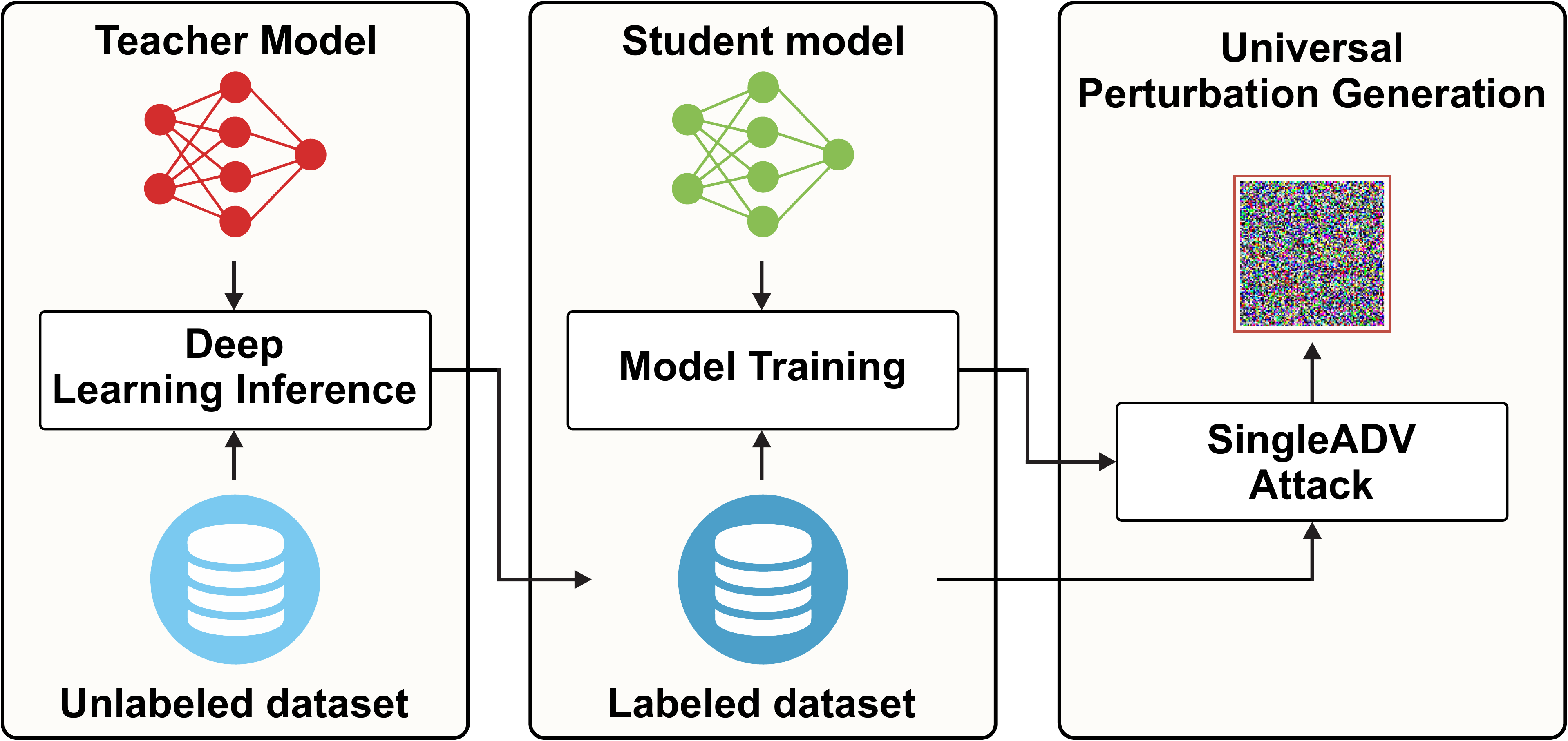}
    \caption{The pipeline of \ours attack in a black-box environment, where the unlabeled dataset is labeled using the teacher model (a black-box model) with the help of a deep learning inference technique. \ours{} uses the student model to learn universal perturbation.}
    \label{fig:blackbox_pipeline}
\end{figure}

\section{\ours{} in Black-box Settings} \label{sec:blackbox}

In \autoref{sec:evaluation}, we conducted experiments on the proposed attack in a white-box environment. Since the white-box attacks have limited practicality in real-life cases, we also employ our proposed method in the black-box settings. In this section, we demonstrate the general applicability of \ours in the black-box settings. The main purpose of this section is to demonstrate the attack usage in a real-world scenario.       

\subsection{Methodology}
As we cannot directly apply \ours to attack a black-box model, we employ one of the techniques against the black-box models \cite{abdukhamidov2022black, cao2021towards}. We use a method called teacher-student learning, considered adequate for attacking DL models with black-box settings \cite{hinton2015distilling}. In teacher-student learning, there is a black-box DL model (referred to as a teacher model) is used to transfer significant knowledge to a DL model (referred to as a student model) \cite{hinton2015distilling}. Specifically, we train a student model based on the output of the teacher model. A well-trained student model can imitate the behavior of the teacher model. Thus, it can be guaranteed that adversarial examples generated against the student model can be directly transferred to the black-box model (teacher model) \cite{cao2021towards}. \autoref{fig:blackbox_pipeline} displays the overall pipeline of \ours in a black-box environment. In the beginning, we prepare a dataset that is unlabeled and relevant to the target black-box deep learning model (teacher model) task. We utilize a technique known as deep learning inference to feed the unlabeled dataset into the teacher model to label the dataset so that a student model can be trained to mimic a teacher model. Once the inference process is finished, we select an architecture for the student model and train it with the labeled dataset. As we mentioned, the architecture of the student model should be complex enough to copy the behavior of the teacher model \cite{cao2021towards}. After training the student model, we adopt it for \ours to generate a universal perturbation based on the selected source and target categories. The \ours generates a single perturbation that triggers the student model to misclassify a specific category. Later, we transfer the induced perturbation to attack the teacher model and test whether the universal perturbation is valid.  

\subsection{Experimental Settings}
\BfPara{Dataset} In the experiment, we randomly select a deep learning app from Google Play to use its deep learning model as a target black-box model. We use an app called Bei Ke that is used to identify scenes. By tracking the APIs of the deep learning frameworks in the app using tools called Soot \cite{soot-oss} and FlowDroid \cite{securesoftwareengineering}, we observe the working and invoking processes of the DL framework. The tools help us invoke the DL model of the app to send a sample and receive its response. We utilize the ImageNet dataset as an unlabeled dataset, which we find most relevant to the app task. We query the teacher model (\ie the DL model within the app) to label the ImageNet dataset, which we find to be the most relevant to the app task. This process involves only the training dataset we used in previous experiments, \ie 1,300 training images per class.
This newly labeled dataset is utilized for training the student model. After the training is complete, 30 random samples from the test set are selected to test our approach against one universal perturbation of the target category.


\BfPara{Student Model} In the teacher-student learning approach, a student model with a complex architecture is the correct option to imitate a teacher model well. In other words, the more complex a student model is, the higher the attack success rate is \cite{cao2021towards}. Considering this, we adopt the VGG model architecture (VGG-11 and VGG-16) as a student model. 

\BfPara{Interpreter} We employ the CAM as the attack showed relatively significant results with the interpreter as shown in \autoref{fig:IOU_sample}. However, the target black-box model (teacher model) does not provide interpretability, and we cannot add the CAM as we do not have access. Therefore, we employ only the fooling ratio metric for the teacher model while calculating the fooling ratio and the similarity of attribution maps for the student model. To show the attack's effectiveness in generating similar interpretations with the adversarial samples as the ones for the benign samples, we perform a controlled experiment where we do not have access to the teacher model but generate interpretation maps from the student model. We adopt ResNet-50 as a teacher model and VGG-16 as a student model. 

For the reproducibility of our experiments, our code, data, and models are available at ({\em\url{https://github.com/InfoLab-SKKU/SingleClassADV}}).

\subsection{Experimental Results}
\autoref{fig:blackbox_result} depicts the results of the experiment. As we mentioned earlier, the app does not provide interpretability, and we tested the app in terms of the fooling ratio. In the figure, the fooling ratio and IoU test results are provided. The IoU test results are calculated from the attribution maps of the student models. The attack achieved over 60\% and 70\% fooling ratios, while the perturbation showed over 60\% and 70\% of similarity in producing adversarial interpretation maps, respectively. The same perturbation reaches about 30\% and 50\% in the fooling ratio when used against the teacher model. The experimental result, displayed in \autoref{fig:blackbox_result}, contains the actual black-box model; therefore, we cannot check the interpretation of the model. Considering the case, we performed a controlled experiment to generate interpretation maps on the teacher model to check if they share similarities with benign interpretation maps. \autoref{fig:blackbox_result_controlled} displays the results of the experiment. The results show that the attack achieved about a 60\% fooling ratio and over 60\% IoU score which is more or less the same as the student model. According to the result of the controlled experiment, we can assume that the teacher model used in the first experiment (see \autoref{fig:blackbox_result}) shares a high similarity in interpretation when an interpreter is coupled. The results of both experiments show the effectiveness of \ours against attack black-box models.

\begin{figure*}[t]
    \begin{minipage}{0.48\textwidth}
    \centering
    \captionsetup{justification=justified}
    \includegraphics[width=0.98\linewidth]{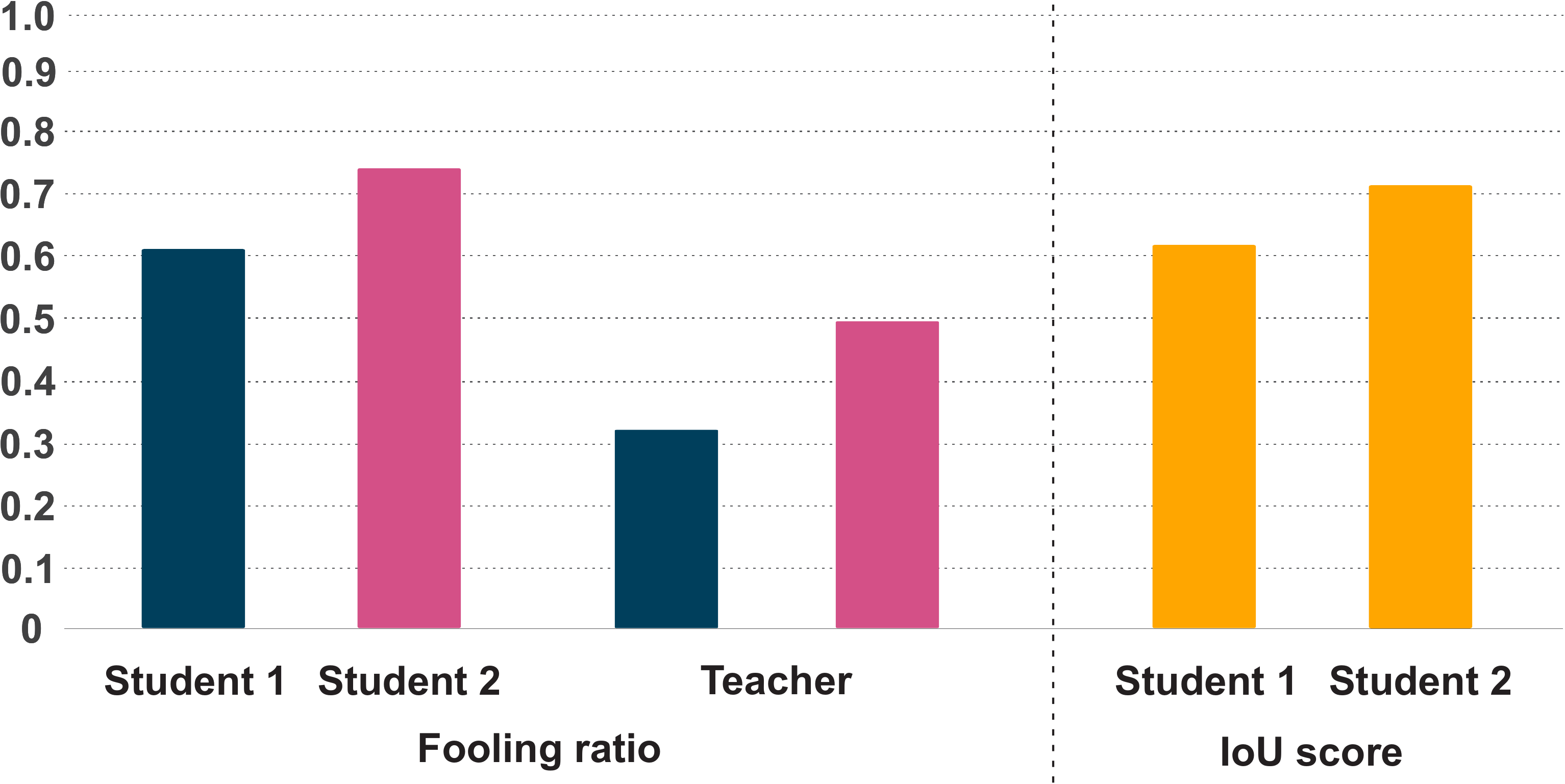}
    \caption{The fooling ratio and IoU scores results are based on the universal perturbation generated by \ours following the teacher-student learning approach. A student refers to a white-box DL model (Student 1: VGG-11, Student 2: VGG-16), and a teacher refers to a black-box DL model used by an app. Source and target categories are selected randomly.}
    \label{fig:blackbox_result}
\end{minipage}~
    \begin{minipage}{0.48\textwidth}
    \centering
    \captionsetup{justification=justified}
    \includegraphics[width=0.72\linewidth]{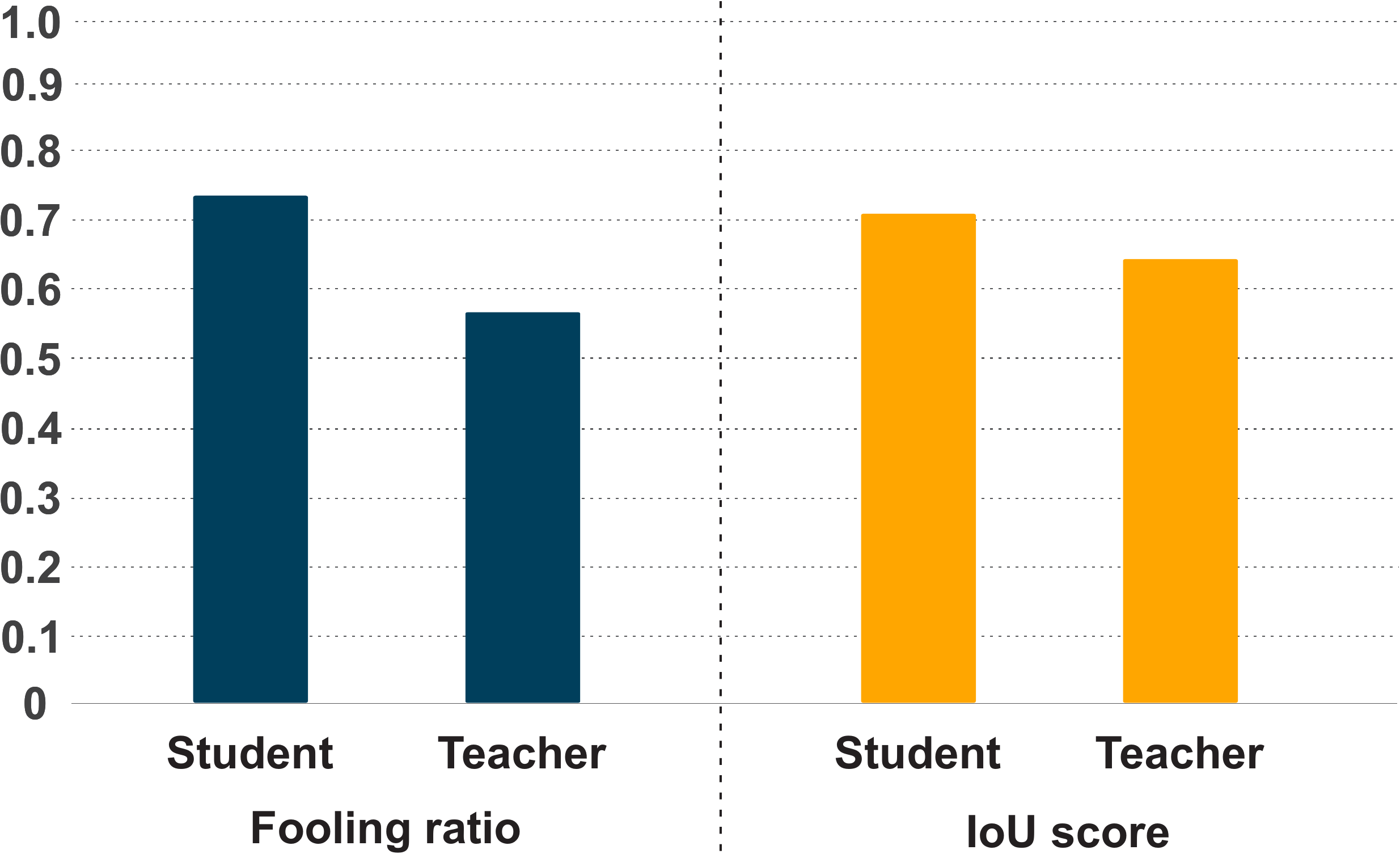}
    \caption{The fooling ratio and IoU scores results are based on the universal perturbation generated by \ours following the teacher-student learning approach. A student refers to a white-box DL model (VGG-16), and a teacher refers to a black-box DL model (ResNet-50). 
    IoU results are based on the attribution maps generated CAM interpreter.}
    \label{fig:blackbox_result_controlled}
    \end{minipage}
\end{figure*}

\section{Countermeasures} \label{sec:countermeasures}
In the following and based on our observations, we discuss several potential countermeasures against \ours. 

\begin{figure*}[t]
    \centering
    \captionsetup{justification=justified}
    \includegraphics[width=0.7\linewidth]{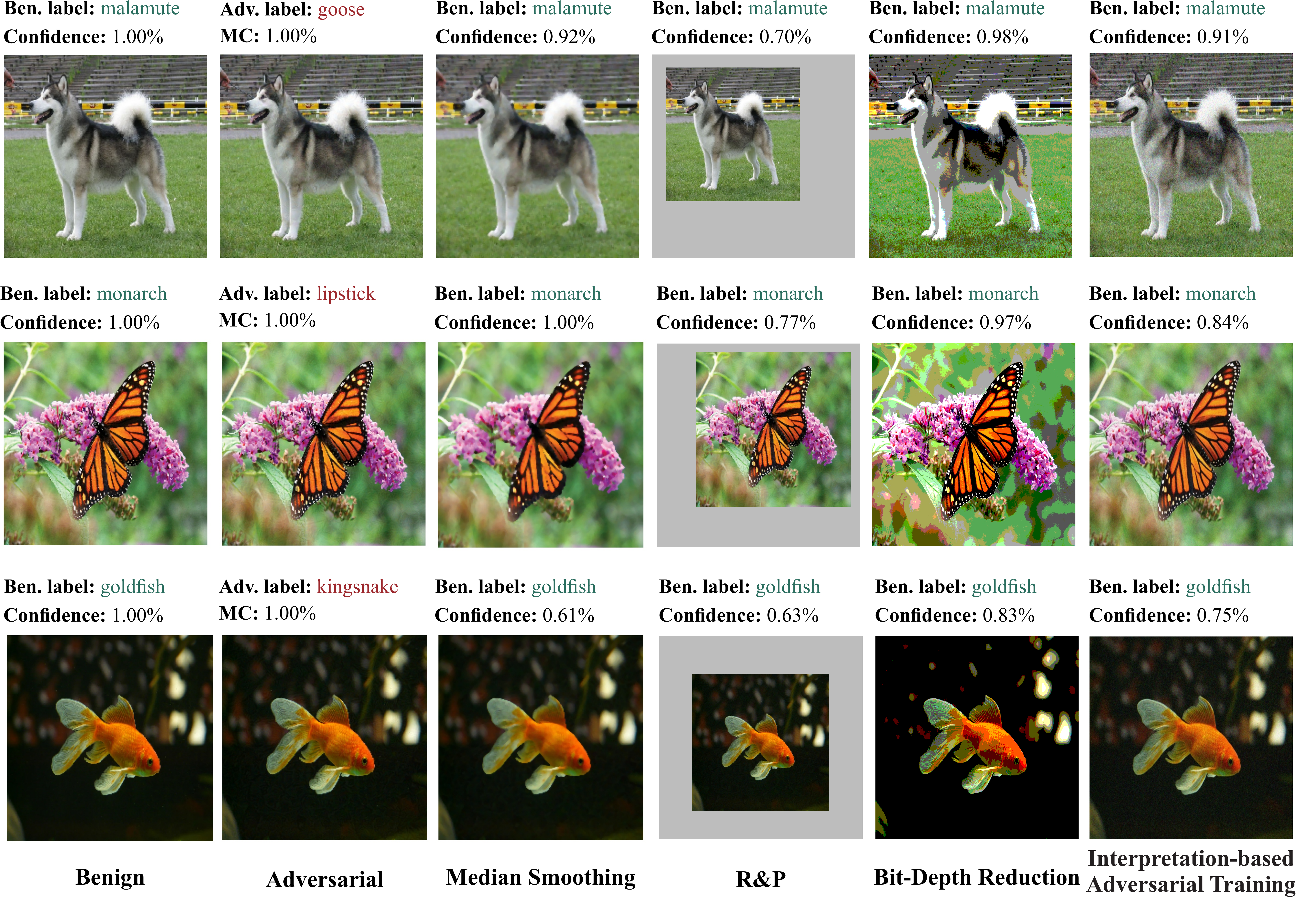}
    \caption{Examples demonstrating the influence of defense mechanisms on adversarial samples. Ben. label and Adv. label represent benign and adversarial labels, respectively. MC stands for misclassification confidence.}
    \label{fig:defense_example}
    \vspace{-2ex}
\end{figure*}

\BfPara{Preprocessing methods} Preprocessing approaches involve specific modifications to eliminate adversarial noise from samples before passing them to the DL model. The objective of preprocessing defenses is to enhance the robustness of DL models against adversarial samples, ensuring that the models can classify adversarial samples correctly with a little reduction in performance on benign images.

It is important to note that relying on a single preprocessing defense technique may not be sufficient, as the attack can be adapted to bypass that particular defense. Therefore, employing multiple preprocessing techniques can help in removing the perturbation added to the samples. By focusing on different features of the samples, these defense techniques make it more challenging to generate robust adversarial samples that can bypass these defenses.

In order to evaluate the effectiveness of defense techniques, including our proposed defense method, we apply them to adversarial samples and visualize their impact in \autoref{fig:defense_example}. These examples provide insights into how the defense models affect the adversarial samples and highlight the effectiveness of the defense techniques in mitigating the impact of \ours. 

\autoref{tab:defense} provides the fooling ratio of \ours when a pair of defense methods is applied to preprocess the adversarial samples generated by \ours. In the experiment, three defense techniques were included, namely bit depth reduction \cite{guo2017countering}, median smoothing \cite{ding2019advertorch}, and random resizing and padding (R\&P) \cite{xie2017mitigating}. Each defense technique was applied with its default hyperparameters. The results demonstrate that employing two defense techniques together decreases the effectiveness of the attack. It is worth noting that the performance of the defense methods can be further improved by adjusting the settings and hyperparameters of the defense techniques. We surmise that optimizing the parameters makes it possible to enhance the performance of the defense methods in mitigating the impact of \ours.


\begin{table}[t]
\centering
\caption{Fooling ratio of adversarial samples when two defense techniques are applied. The results are based on 30 samples of the targeted category (Target 2: {Dog} $\rightarrow$ \textbf{Goose}).}
\arrayrulecolor{black}
\label{tab:defense}
\resizebox{0.9\linewidth}{!}{%
\begin{tabular}{cc}
\toprule
\rowcolor[rgb]{0.718,0.718,0.718} \textbf{Pair of Defenses} & \textbf{Fooling Ratio}  \\ 
\midrule
Bit-Depth Reduction - Median Smoothing                              & 0.13                    \\ 
\midrule
Median Smoothing - R\&P                                        & 0.07                    \\ 
\midrule
R\&P - Bit-Depth Reduction                                           & 0.10                    \\
\bottomrule
\end{tabular}
}
\arrayrulecolor{black}
\end{table}

\BfPara{Ensemble Interpreter} The term ``ensemble interpreter'' refers to using a set of interpreters to provide a comprehensive view of a DL model. By employing multiple interpreters, each with its unique characteristics, a more comprehensive understanding of the DL model can be achieved \cite{10.1007/978-3-030-91434-9_9}. Therefore adopting several interpreters can help detect if a sample is benign or adversarial. 
Future research direction can be in examining whether \ours can be adapted to counter IDLSes with ensemble interpreters. For example, the attack can be optimized to minimize the interpretation loss across multiple interpreters used by an IDLS. However, it is worth mentioning that the generation process of adversarial samples in \ours can be computationally expensive against IDLSes with ensemble interpreters. Therefore, the number of interpreters employed in an IDLS can play a significant role in defending against \ours.



\BfPara{Interpretation-based Adversarial Training} 
The objective of this task is to develop robust classifiers that can effectively handle universal perturbations. To this end, we adopt a similar approach to the one in \cite{madry2017towards} and leverage concepts from robust optimization. Our strategy involves formulating the problem of universal adversarial training as a min-max optimization problem to construct highly resilient models to universal perturbations. By treating the generation of universal perturbations as an optimization task, we aim to find the optimal perturbation that maximally affects the model's robustness while minimizing its impact on the model's performance on benign inputs. To solve this optimization problem, we utilize alternating stochastic gradient methods. 

\begin{algorithm}[t]
\small
\caption{Interpretation-based adversarial training against universal perturbations}\label{alg:adv_training}
\KwData{Samples $X$, perturbation threshold $\epsilon$, learning rate $\alpha$, momentum $\mu$, universal perturbation $\delta$, interpretation masks $M$} 
 
 \For{epoch i = 1...N}{
    \For{minibatch B $\subset$ X and I $\subset$ M}{
    $M_o = \text{\scriptsize \em Binarize}(I, \text{\em t})$\\  
    \textcolor{gray}{\# \em weight update based on expected norms $\mathbb{E}$ of the gradient $\nabla$ } \\
    $g_w \gets \mu g_w - \mathbb{E}_{\substack{x \in B \ m \in M_o}}[\nabla_w L(w, x + m\odot\delta)]$\;
    $w \gets w + \alpha g_w$\;
    \textcolor{gray}{\# \em perturbation update} \\
    $\delta \gets \delta + \epsilon ~ sign(\mathbb{E}_{\substack{x \in B \ m \in M_o}} [\nabla_{\delta} L(w, x + m \odot \delta)])$\;
    Projecting $\delta$ onto the $p$-ball. \\
    } 
 }
\end{algorithm}

\autoref{alg:adv_training} uses a single perturbation refined throughout all iterations. 
We only update the weights $w$ and perturbations $\delta$ once per training step. 
In the algorithm, the universal perturbation is cropped by identifying significant areas. 
This is achieved by converting the interpretation mask $m \in M$ to a binary form $m_o = \text{\em Binarize}(m, \text{\em t})$ using $\text{\em t}$ {(\ie 0.3 for our experiment)} so that a value of $0$ indicating an irrelevant area and $1$ indicating a relevant area. 
The resulting mask is then multiplied with the universal perturbation $\delta$ using element-wise multiplication $\odot$. 
This approach enables the model to learn and become more robust to the perturbation within the interpretation mask. 

For the CIFAR-10 experiment, we utilized a perturbation threshold $\epsilon$ of {0.031}, a batch size of 128, {Momentum SGD with an initial learning rate of 0.1 that drops until 0.001 and trained for 500 epochs} for the {ResNet-20} model. 
The selection of the CIFAR-10 dataset and the ResNet-20 model for adversarial training and testing was driven by practical considerations such as {performance, computational efficiency, standardization, and generalization properties.} 
For the ImageNet experiment, we used pre-calculated universal perturbations (\ie using the DenseNet-169 model with CAM interpreter from the experiments presented in \autoref{tab:fooling_ratio}), as a computationally efficient alternative. 
To enable adversarial training, we employed a fine-tuning approach to learning relevant features in the presence of perturbations efficiently. 
The fine-tuning process utilized a batch size of 32, {Momentum SGD with an initial learning rate of 0.1 that drops until 0.001, a perturbation threshold of 0.04}
and 100 epochs. 

\begin{table}
\centering
\caption{Fooling ratios of \ours using CIFAR-10 and ImageNet datasets before and after interpretation-based adversarial training. The results were obtained using a sample size of 200 per category (2,000 samples in total) for CIFAR-10 and 50 per category (50,000 samples in total) for ImageNet.}
\label{tab:fooling_ratio_adv_training}
\resizebox{0.8\linewidth}{!}{%
\begin{tabular}{ccccc} 
\toprule
\rowcolor[rgb]{0.718,0.718,0.718} \textbf{\#} & \textbf{Source} & \textbf{Target} & \begin{tabular}[c]{@{}>{\cellcolor[rgb]{0.718,0.718,0.718}}c@{}}\textbf{Fooling Ratio~}\\\textbf{(Before)}\end{tabular} & \begin{tabular}[c]{@{}>{\cellcolor[rgb]{0.718,0.718,0.718}}c@{}}\textbf{Fooling Ratio~}\\\textbf{(After)}\end{tabular}  \\ 
\midrule
\multicolumn{5}{c}{\textbf{CIFAR-10}}                                                                                                                                                                                                                                                           \\ 
\midrule
\textbf{1}                                    & Bird            & Airplane        & 0.77                                                                                                                    & 0.15                                                                                                                    \\ 

\textbf{2}                                    & Deer            & Frog            & 0.84                                                                                                                    & 0.16                                                                                                                    \\ 

\textbf{3}                                    & Frog            & Cat             & 0.87                                                                                                                    & 0.17                                                                                                                    \\ 

\textbf{4}                                    & Ship            & Cat             & 0.83                                                                                                                    & 0.16                                                                                                                    \\ 

\textbf{5}                                    & Truck           & Horse           & 0.78                                                                                                                    & 0.15                                                                                                                    \\ 

\textbf{6}                                    & Airplane        & Deer            & 0.80                                                                                                                    & 0.16                                                                                                                    \\ 

\textbf{7}                                    & Horse           & Dog             & 0.84                                                                                                                    & 0.16                                                                                                                    \\ 

\textbf{8}                                    & Dog             & Frog            & 0.69                                                                                                                    & 0.13                                                                                                                    \\ 
\midrule
\multicolumn{5}{c}{\textbf{ImageNet}}                                                                                                                                                                                                                                                           \\ 
\midrule
\textbf{1}                                    & Panda           & Cat             & 0.78                                                                                                                    & 0.14                                                                                                                    \\ 

\textbf{2}                                    & Dog             & Goose           & 0.75                                                                                                                    & 0.15                                                                                                                    \\ 

\textbf{3}                                    & Cup             & Wolf            & 0.69                                                                                                                    & 0.14                                                                                                                    \\
\bottomrule
\end{tabular}
}
\end{table}

In our experiments, we randomly selected the source and target categories for the CIFAR-10 dataset. However, when using the ImageNet dataset, we utilized a previously calculated universal perturbation as described in \autoref{tab:fooling_ratio}. Due to the computational expense of generating a new universal perturbation for large datasets, we opted to use the existing perturbation for our experiments with ImageNet.

To evaluate the effectiveness of interpretation-based adversarial training, we conducted experiments and measured the fooling ratio before and after applying the adversarial training technique. \autoref{tab:fooling_ratio_adv_training} demonstrate that the fooling ratio significantly decreases after applying interpretation-based adversarial training. This indicates that the customized adversarial training approach improves the classifier's robustness against the attack.


\section{Limitations} \label{sec:limiation}


\autoref{tab:fooling_ratio}, \autoref{fig:int_samples} and \autoref{fig:blackbox_result} show that the proposed technique is efficient and effective. However, \autoref{fig:IOU_sample} highlights a limitation in the similarity between the adversarial interpretation maps generated by the MASK interpreter and their benign counterparts. This can be attributed to the construction of the MASK interpreter as an optimization procedure, where even small amounts of noise can lead to variations in the interpretation maps. Despite this limitation, the interpretation maps generated by the MASK interpreter still possess meaningful information, which can deceive the human-involvement process (e.g., expert-checking), as depicted in \autoref{fig:int_samples}.

Another limitation of the proposed attack is the potential impact of the universal perturbation on the classification accuracy of non-source classes. The attack may interfere with the class predictions of different categories, resulting in misclassifications. While the paper focused on the efficiency of the attack regarding the source and target categories, the impact on non-source classes was not considered. One possible solution to address this limitation is to increase the classification confidence of non-source classes for their actual categories while ensuring that they are not misclassified. This can be a potential direction for future research in mitigating the impact of the attack on non-source classes.


\section{Conclusion} \label{sec:conc}

In this paper, we present \ours{}, a targeted adversarial attack that specifically aims to deceive deep learning models in a single object category while minimizing the impact on other categories in both white-box and black-box settings. The proposed attack exploits interpretations to generate universal perturbations that produce adversarial examples with similar interpretations to benign inputs, making them more difficult to detect. Through extensive experiments, we evaluated the effectiveness of \ours{} on four popular deep learning models (\ie VGG-16, ResNet-50, DenseNet-169, and Inception-V3) and three interpretation models (\ie CAM, Grad, and MASK), demonstrating the vulnerability of interpretable deep learning systems to our proposed attack. We explored existing defense techniques to mitigate the impact of such adversarial attacks. Our experiments show that interpretation-based adversarial training can enhance the robustness of models against interpretation-guided attacks.


\BfPara{Future Work} Possible future research directions include investigating the countermeasures and limitations discussed in the paper and exploring the potential applications of the proposed attack and defenses in other domains.
Future studies can also focus on studying the effects of the attack given different model architectures and types of interpretation models.
 \section*{Acknowledgment}
{\small This work was supported by the National Research Foundation of Korea(NRF) grant funded by the Korea government(MSIT)(No. 2021R1A2C1011198), (Institute for Information \& communications Technology Planning \& Evaluation) (IITP) grant funded by the Korea government (MSIT) under the ICT Creative Consilience Program (IITP-2021-2020-0-01821), and AI Platform to Fully Adapt and Reflect Privacy-Policy Changes (No. 2022-0-00688).}

\balance
\bibliographystyle{IEEEtran}
\bibliography{ref}

\begin{thebibliography}{10}
\providecommand{\url}[1]{#1}
\csname url@samestyle\endcsname
\providecommand{\newblock}{\relax}
\providecommand{\bibinfo}[2]{#2}
\providecommand{\BIBentrySTDinterwordspacing}{\spaceskip=0pt\relax}
\providecommand{\BIBentryALTinterwordstretchfactor}{4}
\providecommand{\BIBentryALTinterwordspacing}{\spaceskip=\fontdimen2\font plus
\BIBentryALTinterwordstretchfactor\fontdimen3\font minus
  \fontdimen4\font\relax}
\providecommand{\BIBforeignlanguage}[2]{{%
\expandafter\ifx\csname l@#1\endcsname\relax
\typeout{** WARNING: IEEEtran.bst: No hyphenation pattern has been}%
\typeout{** loaded for the language `#1'. Using the pattern for}%
\typeout{** the default language instead.}%
\else
\language=\csname l@#1\endcsname
\fi
#2}}
\providecommand{\BIBdecl}{\relax}
\BIBdecl

\bibitem{ILSVRC15}
O.~Russakovsky, J.~Deng, H.~Su, J.~Krause, S.~Satheesh, S.~Ma, Z.~Huang,
  A.~Karpathy, A.~Khosla, M.~Bernstein, A.~C. Berg, and L.~Fei-Fei, ``{ImageNet
  Large Scale Visual Recognition Challenge},'' \emph{International Journal of
  Computer Vision (IJCV)}, vol. 115, no.~3, pp. 211--252, 2015.

\bibitem{simonyan2014very}
K.~Simonyan and A.~Zisserman, ``Very deep convolutional networks for
  large-scale image recognition,'' \emph{arXiv:1409.1556}, 2014.

\bibitem{he2016deep}
K.~He, X.~Zhang, S.~Ren, and J.~Sun, ``Deep residual learning for image
  recognition,'' in \emph{Proceedings of the IEEE conference on computer vision
  and pattern recognition}, 2016, pp. 770--778.

\bibitem{he2020towards}
Y.~He, G.~Meng, K.~Chen, X.~Hu, and J.~He, ``Towards security threats of deep
  learning systems: A survey,'' \emph{IEEE Transactions on Software
  Engineering}, 2020.

\bibitem{kindermans2019reliability}
P.-J. Kindermans, S.~Hooker, J.~Adebayo, M.~Alber, K.~T. Sch{\"u}tt,
  S.~D{\"a}hne, D.~Erhan, and B.~Kim, ``The (un) reliability of saliency
  methods,'' in \emph{Explainable AI: Interpreting, Explaining and Visualizing
  Deep Learning}.\hskip 1em plus 0.5em minus 0.4em\relax Springer, 2019, pp.
  267--280.

\bibitem{zhou2016learning}
B.~Zhou, A.~Khosla, A.~Lapedriza, A.~Oliva, and A.~Torralba, ``Learning deep
  features for discriminative localization,'' in \emph{Proceedings of the IEEE
  conference on computer vision and pattern recognition}, 2016.

\bibitem{zhang2018interpretable}
Q.~Zhang, Y.~N. Wu, and S.-C. Zhu, ``Interpretable convolutional neural
  networks,'' in \emph{Proceedings of the IEEE Conference on Computer Vision
  and Pattern Recognition}, 2018, pp. 8827--8836.

\bibitem{10.1007/978-3-030-91434-9_9}
E.~Abdukhamidov, M.~Abuhamad, F.~Juraev, E.~Chan-Tin, and T.~AbuHmed,
  ``Advedge: Optimizing adversarial perturbations against interpretable deep
  learning,'' in \emph{Computational Data and Social Networks}, D.~Mohaisen and
  R.~Jin, Eds.\hskip 1em plus 0.5em minus 0.4em\relax Cham: Springer
  International Publishing, 2021, pp. 93--105.

\bibitem{zhang2020interpretable}
X.~Zhang, N.~Wang, H.~Shen, S.~Ji, X.~Luo, and T.~Wang, ``Interpretable deep
  learning under fire,'' in \emph{29th $\{$USENIX$\}$ Security Symposium
  ($\{$USENIX$\}$ Security 20)}, 2020.

\bibitem{akhtar2021explain}
N.~Akhtar, M.~Jalwana, M.~Bennamoun, and A.~S. Mian, ``Attack to fool and
  explain deep networks,'' \emph{IEEE Transactions on Pattern Analysis and
  Machine Intelligence}, pp. 1--1, 2021.

\bibitem{juraev2022depth}
F.~Juraev, E.~Abdukhamidov, M.~Abuhamad, and T.~Abuhmed, ``Depth, breadth, and
  complexity: Ways to attack and defend deep learning models,'' in \emph{The
  17th ACM ASIA Conference on Computer and Communications Security (ACM ASIACCS
  2022)}, 2022.

\bibitem{abdukhamidov2022interpretations}
E.~Abdukhamidov, M.~Abuhamad, S.~S. Woo, E.~Chan-Tin, and T.~Abuhmed,
  ``Interpretations cannot be trusted: Stealthy and effective adversarial
  perturbations against interpretable deep learning,'' \emph{arXiv preprint
  arXiv:2211.15926}, 2022.

\bibitem{moosavi2017universal}
S.-M. Moosavi-Dezfooli, A.~Fawzi, O.~Fawzi, and P.~Frossard, ``Universal
  adversarial perturbations,'' in \emph{Proceedings of the IEEE conference on
  computer vision and pattern recognition}, 2017, pp. 1765--1773.

\bibitem{khrulkov2018art}
V.~Khrulkov and I.~Oseledets, ``Art of singular vectors and universal
  adversarial perturbations,'' in \emph{Proceedings of the IEEE Conference on
  Computer Vision and Pattern Recognition}, 2018, pp. 8562--8570.

\bibitem{hayes2018learning}
J.~Hayes and G.~Danezis, ``Learning universal adversarial perturbations with
  generative models,'' in \emph{2018 IEEE Security and Privacy Workshops
  (SPW)}.\hskip 1em plus 0.5em minus 0.4em\relax IEEE, 2018, pp. 43--49.

\bibitem{heo2019fooling}
J.~Heo, S.~Joo, and T.~Moon, ``Fooling neural network interpretations via
  adversarial model manipulation,'' \emph{Advances in Neural Information
  Processing Systems}, vol.~32, 2019.

\bibitem{ghorbani2019interpretation}
A.~Ghorbani, A.~Abid, and J.~Zou, ``Interpretation of neural networks is
  fragile,'' in \emph{Proceedings of the AAAI conference on artificial
  intelligence}, vol.~33, no.~01, 2019, pp. 3681--3688.

\bibitem{sabour2017dynamic}
S.~Sabour, N.~Frosst, and G.~E. Hinton, ``Dynamic routing between capsules,''
  \emph{arXiv:1710.09829}, 2017.

\bibitem{murdoch2018beyond}
W.~J. Murdoch, P.~J. Liu, and B.~Yu, ``Beyond word importance: Contextual
  decomposition to extract interactions from lstms,'' \emph{arXiv:1801.05453},
  2018.

\bibitem{dabkowski2017real}
P.~Dabkowski and Y.~Gal, ``Real time image saliency for black box
  classifiers,'' \emph{arXiv:1705.07857}, 2017.

\bibitem{simonyan2014deep}
K.~Simonyan, A.~Vedaldi, and A.~Zisserman, ``Deep inside convolutional
  networks: Visualising image classification models and saliency maps,'' in
  \emph{Proceedings of the International Conference on Learning Representations
  ({ICLR})}, 2014.

\bibitem{shrikumar2017learning}
A.~Shrikumar, P.~Greenside, and A.~Kundaje, ``Learning important features
  through propagating activation differences,'' in \emph{International
  conference on machine learning}.\hskip 1em plus 0.5em minus 0.4em\relax PMLR,
  2017, pp. 3145--3153.

\bibitem{smilkov2017smoothgrad}
D.~Smilkov, N.~Thorat, B.~Kim, F.~Vi{\'e}gas, and M.~Wattenberg, ``Smoothgrad:
  removing noise by adding noise,'' \emph{arXiv:1706.03825}, 2017.

\bibitem{selvaraju2017grad}
R.~R. Selvaraju, M.~Cogswell, A.~Das, R.~Vedantam, D.~Parikh, and D.~Batra,
  ``Grad-cam: Visual explanations from deep networks via gradient-based
  localization,'' in \emph{Proceedings of the IEEE international conference on
  computer vision}, 2017, pp. 618--626.

\bibitem{fong2017interpretable}
R.~C. Fong and A.~Vedaldi, ``Interpretable explanations of black boxes by
  meaningful perturbation,'' in \emph{Proceedings of the IEEE international
  conference on computer vision}, 2017, pp. 3429--3437.

\bibitem{abdukhamidov2021advedge}
E.~Abdukhamidov, M.~Abuhamad, F.~Juraev, E.~Chan-Tin, and T.~AbuHmed,
  ``Advedge: Optimizing adversarial perturbations against interpretable deep
  learning,'' in \emph{Computational Data and Social Networks: 10th
  International Conference, CSoNet 2021, Virtual Event, November 15--17, 2021,
  Proceedings 10}.\hskip 1em plus 0.5em minus 0.4em\relax Springer, 2021, pp.
  93--105.

\bibitem{abdukhamidov2022black}
E.~Abdukhamidov, F.~Juraev, M.~Abuhamad, and T.~Abuhmed, ``Black-box and
  target-specific attack against interpretable deep learning systems,'' in
  \emph{Proceedings of the 2022 ACM on Asia Conference on Computer and
  Communications Security}, 2022, pp. 1216--1218.

\bibitem{huang2017densely}
G.~Huang, Z.~Liu, L.~Van Der~Maaten, and K.~Q. Weinberger, ``Densely connected
  convolutional networks,'' in \emph{Proceedings of the IEEE conference on
  computer vision and pattern recognition}, 2017, pp. 4700--4708.

\bibitem{szegedy2016rethinking}
C.~Szegedy, V.~Vanhoucke, S.~Ioffe, J.~Shlens, and Z.~Wojna, ``Rethinking the
  inception architecture for computer vision,'' in \emph{Proceedings of the
  IEEE conference on computer vision and pattern recognition}, 2016, pp.
  2818--2826.

\bibitem{he2017mask}
K.~He, G.~Gkioxari, P.~Doll{\'a}r, and R.~Girshick, ``Mask r-cnn,'' in
  \emph{Proceedings of the IEEE international conference on computer vision},
  2017, pp. 2961--2969.

\bibitem{cao2021towards}
H.~Cao, S.~Li, Y.~Zhou, M.~Fan, X.~Zhao, and Y.~Tang, ``Towards black-box
  attacks on deep learning apps,'' \emph{arXiv:2107.12732}, 2021.

\bibitem{hinton2015distilling}
G.~Hinton, O.~Vinyals, J.~Dean \emph{et~al.}, ``Distilling the knowledge in a
  neural network,'' \emph{arXiv:1503.02531}, vol.~2, no.~7, 2015.

\bibitem{soot-oss}
\BIBentryALTinterwordspacing
Soot-Oss, ``Soot-oss/soot: Soot - a java optimization framework.'' [Online].
  Available: \url{https://github.com/soot-oss/soot}
\BIBentrySTDinterwordspacing

\bibitem{securesoftwareengineering}
\BIBentryALTinterwordspacing
Secure-Software-Engineering, ``Secure-software-engineering/flowdroid: Flowdroid
  static data flow tracker.'' [Online]. Available:
  \url{https://github.com/secure-software-engineering/FlowDroid}
\BIBentrySTDinterwordspacing

\bibitem{guo2017countering}
C.~Guo, M.~Rana, M.~Cisse, and L.~Van Der~Maaten, ``Countering adversarial
  images using input transformations,'' \emph{arXiv:1711.00117}, 2017.

\bibitem{ding2019advertorch}
G.~W. Ding, L.~Wang, and X.~Jin, ``{AdverTorch} v0.1: An adversarial robustness
  toolbox based on pytorch,'' \emph{arXiv:1902.07623}, 2019.

\bibitem{xie2017mitigating}
C.~Xie, J.~Wang, Z.~Zhang, Z.~Ren, and A.~Yuille, ``Mitigating adversarial
  effects through randomization,'' \emph{arXiv:1711.01991}, 2017.

\bibitem{madry2017towards}
A.~Madry, A.~Makelov, L.~Schmidt, D.~Tsipras, and A.~Vladu, ``Towards deep
  learning models resistant to adversarial attacks,'' \emph{arXiv:1706.06083},
  2017.

\end{thebibliography}



\end{document}